\documentclass[12pt]{article}
\usepackage[utf8]{inputenc}

\usepackage[margin=1.00in]{geometry}
\usepackage{graphicx}
\usepackage{caption}
\usepackage{subcaption}
\usepackage{amsmath}
\usepackage{setspace}
\usepackage{booktabs}
\usepackage{xcolor}
\usepackage{scicite}
\usepackage{times}
\usepackage{bbm}
\usepackage{setspace}
\usepackage{titling}
\usepackage{changepage}
\usepackage{upgreek}
\usepackage{bibentry}

\usepackage[hyphens]{url}
\usepackage[hidelinks]{hyperref}

\renewcommand{\figurename}{Fig.}

\newenvironment{sciabstract}{%
\begin{quote} \bf}
{\end{quote}}

\title{Locating and measuring marine aquaculture production from space: a computer vision approach in the French Mediterranean\vspace{-1ex}}

\author
{\normalsize{Sebastian Quaade,$^{1\dag}$ Andrea Vallebueno,$^{1\dag}$ Olivia D. N. Alcabes,$^{1,2}$  Kit T. Rodolfa,$^{1\ddag}$ Daniel E. Ho$^{1\ddag\ast}$}\\
\\
\normalsize{$^{1}$Regulation, Evaluation and Governance Lab,}\\\normalsize{Stanford University, Stanford, California, USA}\\
\normalsize{$^{2}$California Institute of Technology,}\\
\normalsize{Pasadena, California, USA}\\
\\
\normalsize{$^\dag$These authors contributed equally to this work}\\
\normalsize{$^\ddag$Equal co-supervision}\\
\normalsize{$^\ast$To whom correspondence should be addressed; E-mail:  deho@stanford.edu.}
}

\date{}

\begin{document}

%\doublespacing
\baselineskip24pt

\maketitle

\textbf{Short title: } Locating marine aquaculture production from space

\begin{sciabstract}
    Aquaculture production -- the cultivation of aquatic plants and animals -- has grown rapidly since the 1990s, but sparse, self-reported and aggregate production data limits the effective understanding and monitoring of the industry's trends and potential risks. Building on a manual survey of aquaculture production from remote sensing imagery, we train a computer vision model to identify  marine aquaculture cages from aerial and satellite imagery, and generate a spatially explicit dataset of finfish  production locations in the French Mediterranean from 2000-2021 that includes 4,010 cages (69$m^2$ average  cage area). We demonstrate the value of our method as an easily adaptable, cost-effective approach that can improve the speed and reliability of aquaculture surveys, and enables downstream analyses relevant to researchers and regulators. We illustrate its use to compute independent estimates of production, and develop a flexible framework to quantify uncertainty in these estimates. Overall, our study presents an efficient, scalable and highly adaptable method for monitoring aquaculture production from remote sensing imagery. 
\end{sciabstract}

\textbf{Teaser: }
Object detection methods on remote sensing imagery can be effective tools to monitor aquaculture production around the world.  

\section*{Introduction}

In the last 30 years, global seafood production has surged as a result of growing populations, rising incomes, changing consumer preferences and technological innovations~\cite{fao-2022}. 
Aquaculture -- the cultivation of aquatic plants and animals -- has emerged as a key driver of the supply of aquatic foods. 
With a 6-fold increase in global production between 1990 and 2020 (average growth rate of 6.7\% per year)~\cite{fao-2022}, it is the fastest growing food production sector~\cite{klinger_searching_2012,troell2014does}, outputting 122.6 million tonnes (live weight) in 2020 across freshwater, brackish water and marine culture systems~\cite{fao-2022}.
The overexploitation of wild fish, rising global trade of aquatic foods, and ongoing developments leading to more efficient industry practices 
point to the continued growth of aquaculture and to the strengthening of its role as a key contributor to seafood supply~\cite{naylor_20-year_2021}.

Aquaculture has been touted for its potential to increase food security and nutrition~\cite{gentry_mapping_2017,costello_future_2020,golden2021aquatic}, but has also been met with environmental and animal welfare concerns. In 2018, aquatic animals provided 15.3\% of global crude protein supply -- 7.4\% stemming from aquaculture production --~\cite{boyd2022contribution}, and estimates indicate that aquatic food production could contribute up to 8\% of zinc and iron, and 27\% of vitamin B12 global supply in 2030~\cite{golden2021aquatic}.
A growing ``blue food'' movement argues that aquatic foods can contribute to improved nutrition with less pronounced environmental impacts than other sources of animal protein~\cite{gephart2021environmental}, though benefits to food security and environmental footprint can vary significantly across aquatic food production systems, cultures and practices~\cite{tigchelaar2022vital}.
At the same time, aquaculture is associated with a number of environmental harms. Inefficient feeding and animal waste can effect heightened concentrations of organic matter in water bodies, leading to the development of algal blooms that adversely affect water quality for humans and broader ecosystem health~\cite{dauda2019waste,bureau2010towards,Ren:2014aa,wang_aquaculture_2020}. Densely stocked aquaculture farms also regularly administer antibiotics to control or prevent disease outbreaks, and the industry's use intensity of antimicrobials has been estimated to exceed human and terrestrial animal consumption levels~\cite{schar_global_2020}. 
A review of studies examining the use of antibiotics in aquaculture across 15 major producing countries suggests the industry's antibiotic use
has contributed to the emergence of at least 12 multi-antimicrobial resistant (MAMR) pathogen strains, which impose higher mortality rates and costs on aquaculture~\cite{lulijwa_antibiotic_2020}. These resistances can also spread to bacteria that infect humans~\cite{pepi2021antibiotic}. Additionally, several scholars have raised concerns regarding the welfare of intensively farmed aquatic animals~\cite{ashley_fish_2007,conte_stress_2004,macquarie_university_pain_2019}, as few are adapted to conditions commonly found in intensive aquaculture operations~\cite{franks_animal_2021}.

Aquaculture's rapid growth and environmental footprint has increased demand for timely and reliable data on the industry. The Food and Agriculture Organization (FAO) generates the only source of data on fisheries and aquaculture production at a global level~\cite{fao-2022}. On an annual basis, it publishes country-level aquaculture production data, which are primarily assembled from surveys conducted by national statistical offices~\cite{fao2023aquaculture}. While FAO leverages alternative data sources -- such as reports from regional fishery bodies -- to compile these monitoring statistics as accurately as possible, their reliance on national reports implies that the quality and availability of data is highly related to that of the surveys. In addition to general non-reporting, these can suffer from partial information, inconsistencies, highly aggregated data that fail to meet reporting standards, and varying data quality across species and production systems~\cite{fao-2022}. In fact, leading aquaculture countries have in the past been found to misreport production statistics~\cite{campbell2013mariculture}.
In 2020, the FAO indicated that only 59\% of countries submitted or reported official aquaculture data -- out of a total of 207 producing countries and territories -- although these represented 97.6\% of global production~\cite{fao-2022}.   

Spatially granular and disaggregated data on aquaculture production is key to effective governance, monitoring and regulation. Many of aquaculture's negative environmental effects  -- including nutrification, chemical pollution and threats to marine mammals~\cite{espinosa-miranda_entanglements_2020,herediaazuaje_pinnipeds_2022,harnish_longterm_2023} -- are highly localized, and its environmental impact varies across production systems~\cite{tigchelaar2022vital}. 
Production primarily includes inland aquaculture (44\% of global production in 2020) -- which takes place in freshwater systems located in waterways such as rivers, ponds or canals -- and marine aquaculture (``mariculture" - 47\% of production) -- which cultivates species at sea during the entirety of the life cycle or exclusively during the grow-out stage~\cite{fao_fao_2020,fao-2022}. 
In particular, few sources of data specify the location and intensity of mariculture  with high spatial resolution.  The European Marine Observation and Data Network (EMODNet) compiles data on marine finfish production locations using government reports from European countries~\cite{european_marine_observation_and_data_network_emodnet_emodnet_2021}. However, only Cyprus, Denmark, Finland, Greece, Ireland, Malta, Norway, Spain and Scotland report such data. More ambitiously, Clawson and colleagues \cite{CLAWSON2022738066} derive a spatially detailed dataset of global mariculture production by aggregating aquaculture data from national data releases, peer reviewed research, and data releases from industry and research organizations. Although they are able to allocate 96\% of mariculture production to specific spatial locations, only 17\% of farms in their dataset have known locations, while the locations of the remaining 83\% are imputed using information on suitable farm locations. Furthermore, at a 25km$^2$ resolution, the dataset may not be sufficiently granular to assess the effects of mariculture effluents that are spatially concentrated. To derive higher resolution distributions of mariculture production across space, which is critical to monitoring risks to production and the environment, more data is needed on the location and size of these farms.

One way to obtain precise marine aquaculture locations is to identify farms from remote sensing imagery. For instance, \cite{trujillo_fish_2012} and \cite{katselis2022mapping} manually scan the Mediterranean and Greek coasts, respectively, using Google Earth Pro to develop bottom-up inventories of marine aquaculture production locations. Manual labeling efforts like these can provide valuable detailed information. At the same time, they are highly time consuming, and do not easily replicate or scale over time and space. A large literature uses a variety of classical spectral, spatial features analysis, and object-based image analysis methods to classify aquaculture production areas in satellite imagery 
(see for instance \cite{ottinger2018opportunities} and \cite{zhang2013extraction}). However, many of these methods rely on a small number of tunable parameters, meaning they can struggle to achieve good fit when aquaculture facilities and ocean backgrounds vary significantly in appearance~\cite{Liu:2019aa}.

More recent work takes advantage of developments in computer vision to improve the accuracy of automated methods for mapping food production from remote sensing imagery. In the United States, deep learning approaches, namely object detection and segmentation models, have been used to locate land-based agriculture facilities that haven't been previously reported~\cite{Handan-Nader:2019aa,Robinson2022}. Several studies have also used deep learning to map mariculture in China. For example, a variety of image segmentation models have been trained to recognize aquaculture areas in low to medium resolution images, ranging from specific regional maps to mapping all mariculture facilities along the Chinese coast~\cite{Shi2018,fu2021new,zou2022extraction,su2022raftnet}. Other researchers have trained image segmentation models to map mariculture farms on high resolution imagery, although they have resorted to small study areas due to limited data availability~\cite{fu2019finer,Liu:2019aa}.
These studies have been primarily methodological in nature, designing their own deep learning architectures to enhance models' capacity to  segment aquaculture rafts. The literature has underexplored whether off-the-shelf deep learning methods -- such as widely available, pre-trained object detection models -- can be readily deployed by researchers to monitor aquaculture production with reasonable performance. Moreover, to our knowledge, deep learning-based approaches for identifying marine finfish production have not been developed for other regions around the world. 

To address the scarcity of high resolution geospatial mariculture data, we develop a computer vision based method for identifying marine finfish farms from remote sensing imagery. 
Building on a manual mapping of marine finfish cage locations in the Mediterranean by \cite{trujillo_fish_2012}, we first create a hand-labeled dataset of mariculture finfish cages on high-resolution satellite imagery from Google Earth Pro. Next, we fine-tune a YOLOv5 object detection model on this dataset to identify individual cages in remote sensing images, and develop an effective cage detection post-processing procedure to further enhance model performance. 
We evaluate our model's performance and demonstrate the value of our method for aquaculture monitoring in a representative setting -- the entire French Mediterranean coast -- using publicly available aerial imagery. To illustrate the impactful, downstream analyses that can be performed with our method, we create an inventory of marine finfish facilities and facility size in this region over time. Moreover, we generate production estimates for the region -- with accompanying uncertainty measures -- and compare these temporal estimates to survey-based data from the FAO.

Our work makes several contributions to the growing literature applying deep learning techniques to aquaculture mapping.
First, we provide a methodology for offshore aquaculture detection that is adaptable, cost-effective, easy to implement, and requires small amounts of training data. This offers aquaculture researchers a highly attractive alternative for conducting aquaculture surveys and monitoring industry trends over time -- such as facilities' location and size -- compared to manual scans of coastal imagery and self-reported statistics. Moreover, our method's use of a pre-trained object detection model paired with high resolution remote sensing imagery underpins its simplicity relative to other computer vision based applications of aquaculture mapping.
Second, we develop and demonstrate a procedure for estimating marine finfish production, providing an alternate source of information that can be used to verify other production data. Our highly flexible production estimation approach is designed to capture uncertainty in our estimates stemming from multiple sources, including model performance, idiosyncrasies in the availability of aerial imagery, and uncertainty in the distribution of the factors of production. This approach to measuring uncertainty is not only valuable for aquaculture estimation efforts, but for machine learning based methods in general seeking to compute production estimates via remote sensing.
Third, we showcase how our model can be used as part of a human-AI collaborative system, by identifying candidate aquaculture locations for human review. On a standalone basis, 
our model recovers 73\% (79\%) of hand-annotated cages (cage clusters) from an independent test set of aerial imagery along the French Mediterranean coast, illustrating the potential of a human-AI system to improve cage-level performance and conduct more efficient surveys of aquaculture production. By manually annotating fewer than 3\% of all the French Mediterranean coastal imagery from 2000-2021, this human-in-the-loop approach is able to capture all detected cage clusters, and  produce a robust upper bound on the population of cages in the region. 
Fourth, we release a dataset of aerial images of the French Mediterranean -- and a dataset of satellite images from the Mediterranean that is available to other academic researchers upon request -- that can be used by researchers to implement our method in other producing regions. In combination, these datasets contain 6,481 square and circular surface cage bounding box annotations. 
To our knowledge, our work represents the first deep-learning based mariculture mapping application outside of China. 

\begin{figure}[t]
    \centering
    \includegraphics[trim={2cm 0cm 2cm 0cm}, clip,width=0.35\columnwidth]{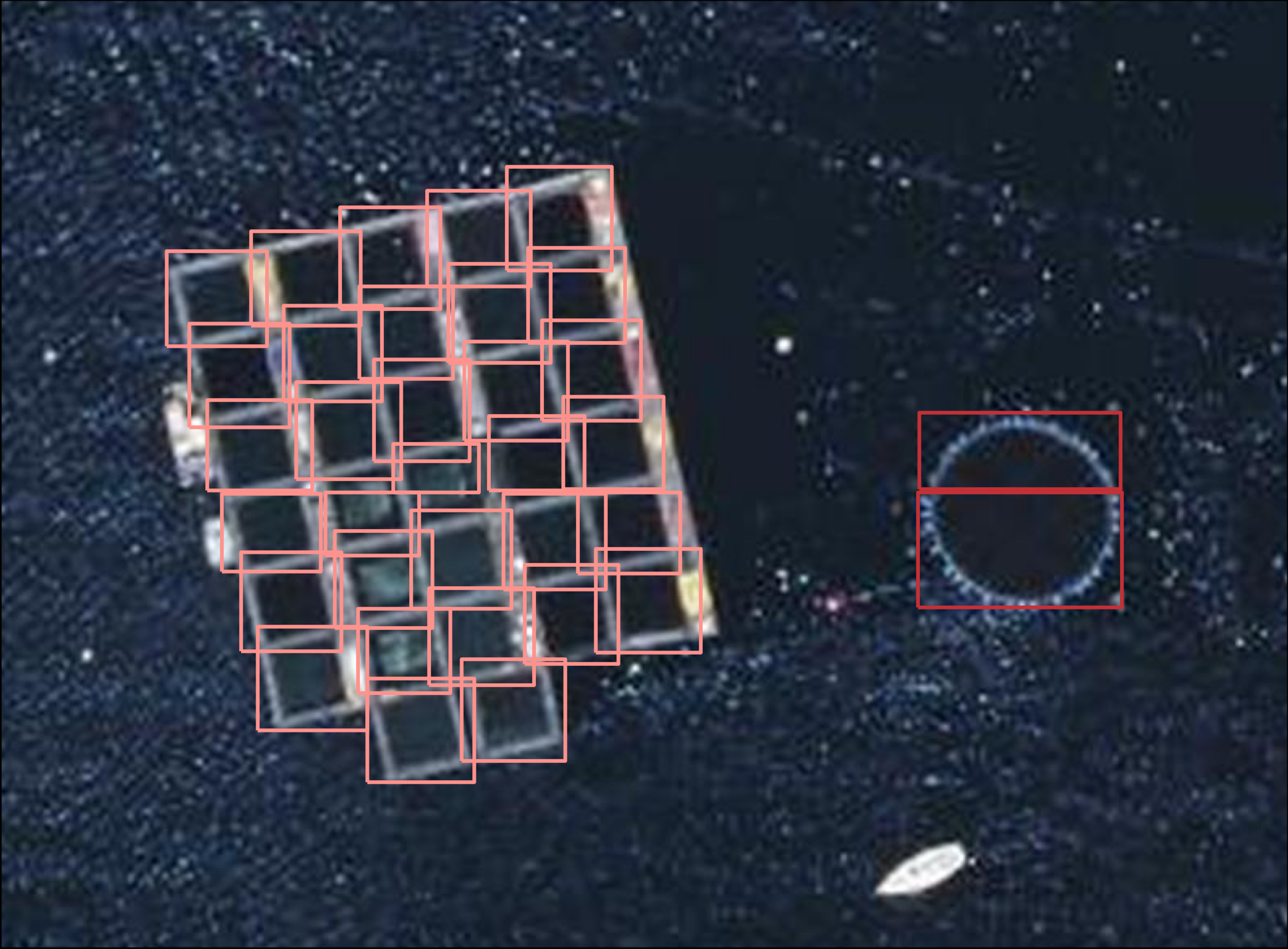}
    \includegraphics[trim={0cm 0cm 0cm 1.8cm}, clip,width=0.35\columnwidth]{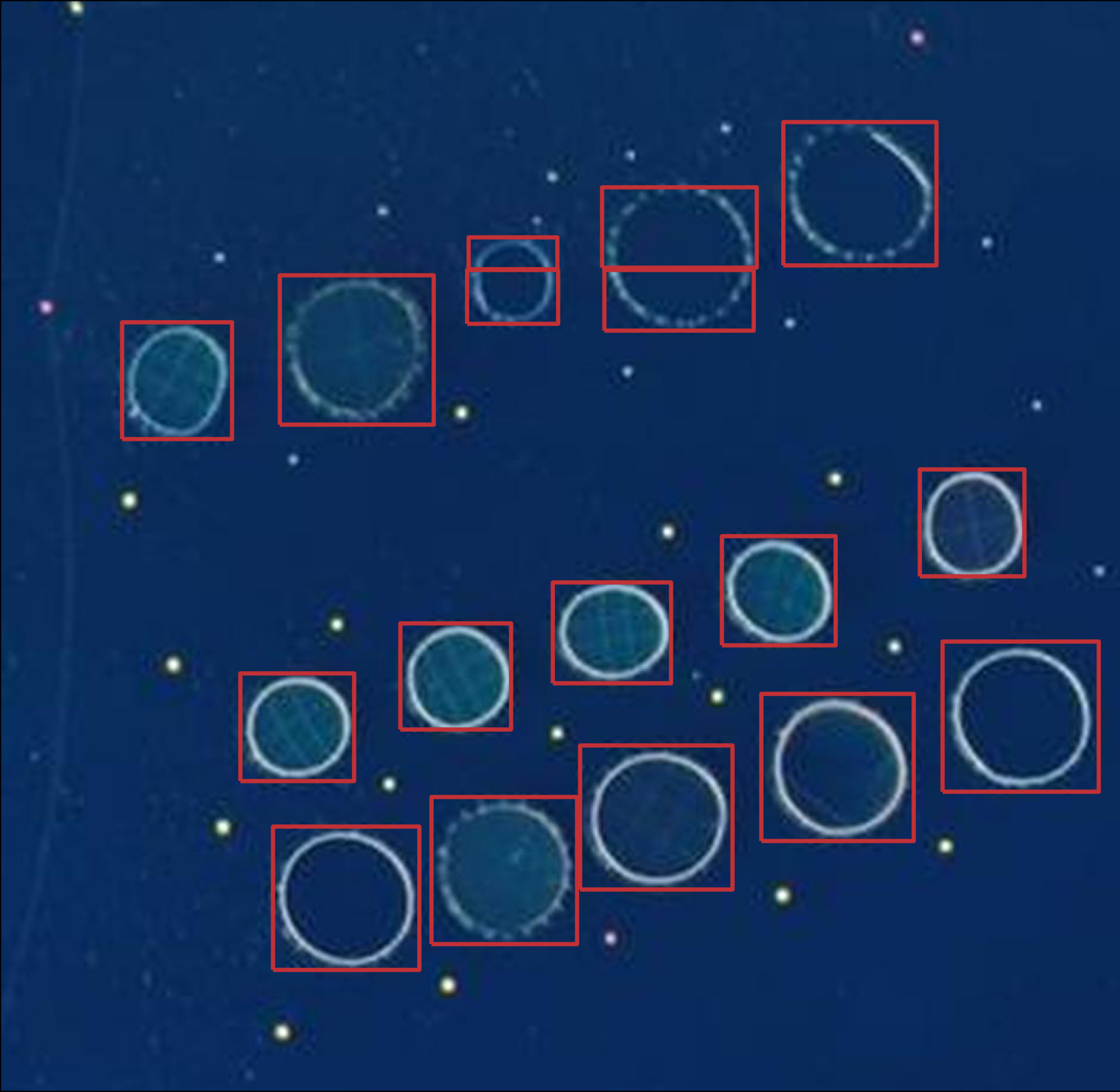}
    \caption{\textbf{Example detections of  aquaculture cages.} Detections of surface square (pink) and circular (red) aquaculture cages identified by our object detection model on aerial imagery of the French Mediterranean. Imagery: Institut national de l'information géographique et forestière (IGN)}
    \label{fig:inference_examples}
\end{figure}
 
\section*{Results} \label{sec:results}

\subsection*{A Computer Vision Model to Detect Aquaculture Cages}
Offshore finfish aquaculture primarily utilizes floating cage structures, which are often placed in clusters near feed storage, operation platforms and other central resources. As the cage structures belong to two main typologies (square or circular) with differing geometries, we opted for a computer vision model that could not only detect the cage instances, but identify their typology. This enables downstream calculations of the cage area using the detected bounding boxes, and thus, the estimation of production tonnage. 
\figurename~\ref{fig:inference_examples} illustrates these different types of cage structures and showcases our model's ability to detect and distinguish the two typologies.

\begin{figure}[ht!]
    \centering
    \includegraphics[width=0.8\columnwidth]{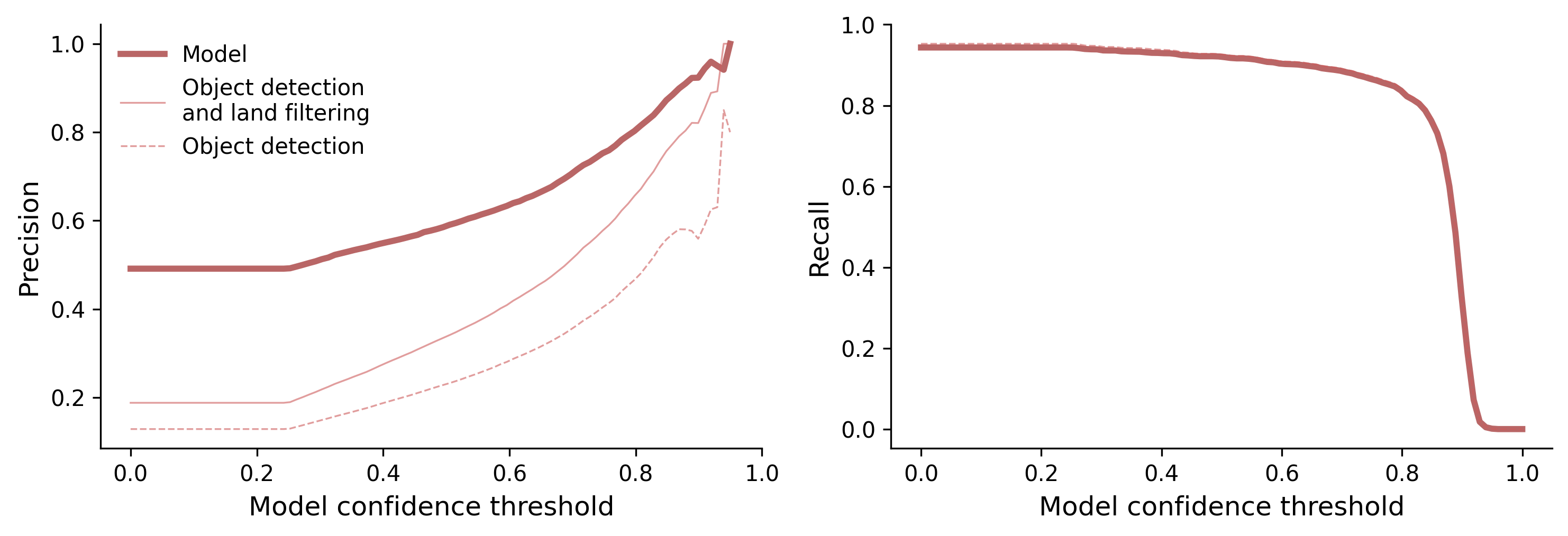}
    \caption{\textbf{Performance on the French Mediterranean coast.} We measure our method's performance in terms of precision and recall  at the cage level. 
    The dark line reflects performance of the overall methodology (\textit{i.e.}, the object detection model in addition to the removal of land-based detections, and cage clustering), whereas lighter lines reflect the standalone performance of the object detection model without these sequential post-processing steps. Recall is virtually unchanged across the detection model and post-processing steps. 
    }
    \label{fig:inference_performance}
\end{figure}

We evaluate our method's performance along the entire French Mediterranean coast -- a representative setting of a region of interest for aquaculture activity detection -- on aerial imagery from 2000 to 2021 using the sampling approach described in the Methods section.  
Our overall methodology involves obtaining predictions from an object detection model, followed by two post-processing steps. 
The first post-processing step removes land-based objects by using a geometry of the French coastline to filter out all detections that are not located in the ocean. Next, we use the DBSCAN clustering algorithm to aggregate the cage-level predictions into cage groups similar to the structures that are typically operated by aquaculture facilities. As we are unable to distinguish aquaculture facilities -- ownership of the cage groups cannot be determined solely from the aerial imagery -- we refer to these sets of cages as clusters rather than facilities. This clustering is based on cage proximity and the number of cages within the cage cluster, and is an effective approach to remove isolated predictions, which are often false positive. 
\figurename~\ref{fig:inference_performance} visualizes our method's performance across model confidence scores on the complete coastal imagery of the region. At a confidence threshold of 0.80, it achieves cage-level precision and recall of 82\%. Moreover, this figure illustrates the considerable value of the methodology's post-processing steps for overall precision, particularly when compared to the standalone object detection model (see the Supplementary Materials for typical false positive predictions; \figurename~\ref{fig:inference_false_positives}).  
We tuned the post-processing procedure via a 5-fold cross validation approach using the hyperparameter grid search procedure described in the Supplementary Materials. We found the hyperparameters that maximized the product of recall and precision on the folds to be a model score threshold of 0.785, a distance threshold of 50 meters, and a minimum cluster size of 5 cage predictions. These parameters achieve 91\% precision and 73\% recall at the cage-level, and 82\% precision and 79\% recall at the cage cluster-level, on an independent test set comprising 10\% of the region's aerial imagery. 

\begin{figure}[t!]
    \centering
    \includegraphics[ width=0.6\columnwidth]{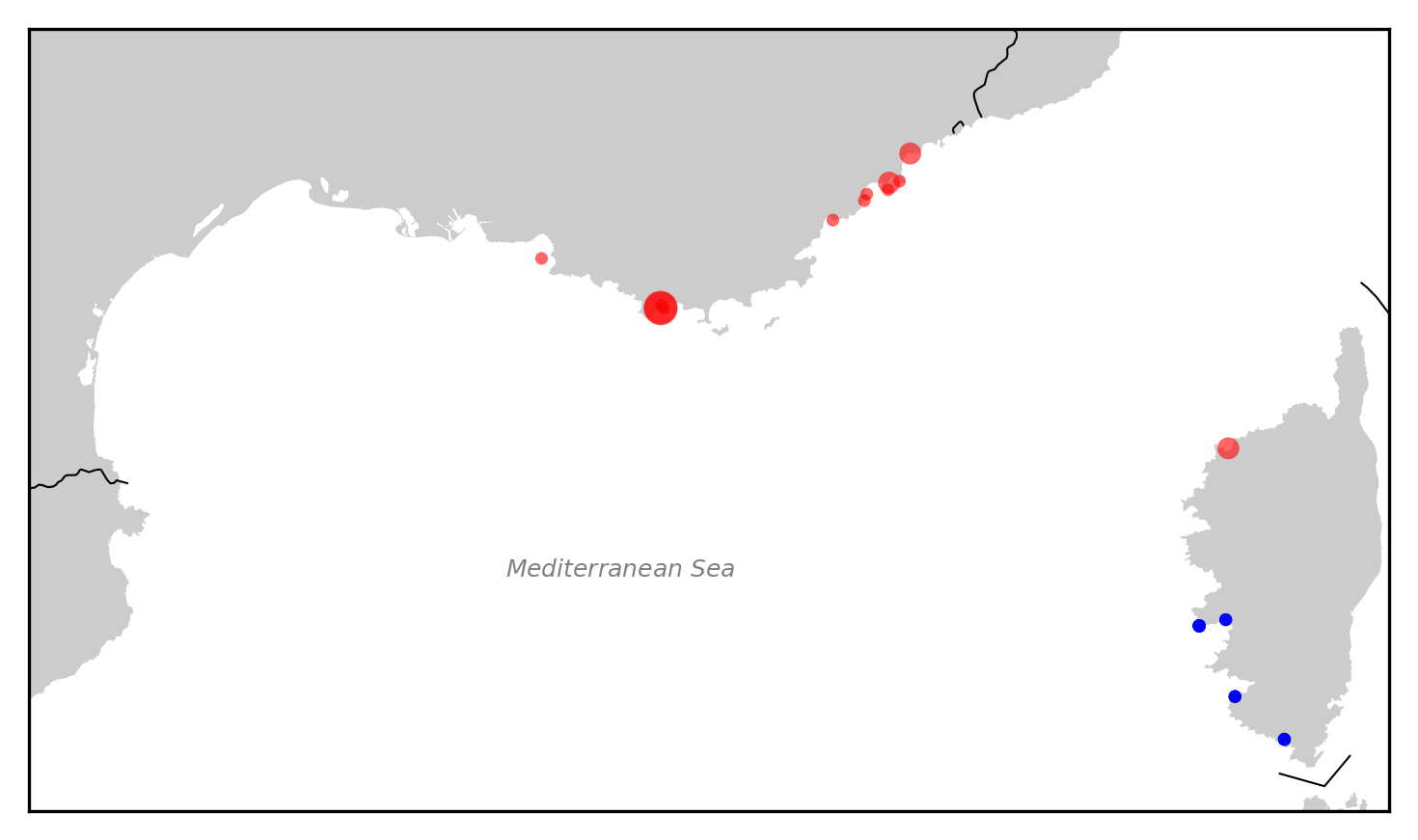}
    \includegraphics[width=0.5\columnwidth]{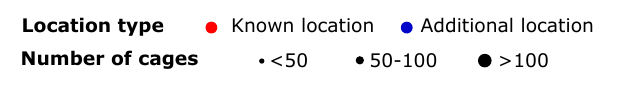} \\
    \caption{\textbf{Marine finfish aquaculture production locations in the French Mediterranean.}  Red points indicate the known locations found by \cite{trujillo_fish_2012} in their manual survey of 2002-2010 Google Earth. 
    Blue points indicate cage clusters detected by our model during 2000-2021 that are at least one kilometer away from these known locations.}
    \label{fig:french_facilities}
\end{figure}

Our methodology's ability to effectively locate aquaculture production is underlined by its performance relative to manual labeling efforts. 
Due to the cadence of the French Mediterranean coastal imagery, there is a slight mismatch between the study period of Trujillo et al.'s~\cite{trujillo_fish_2012} manual survey of the region (performed from 2002 to 2010) and the timing of our imagery. Nonetheless, we make the closest comparison possible by considering cage clusters from 2000-2004, 2005-2009 and 2010-2012 to find new locations within \cite{trujillo_fish_2012}'s time period, and cage clusters from 2013-2015, 2016-2018 and 2019-2021 to find new locations that postdate their study period.
In this manner, we find all of the marine finfish aquaculture locations in the French Mediterranean that were identified by \cite{trujillo_fish_2012} in their manual survey of Google Earth from 2002 to 2010. During this study period, our model also finds an additional 7 cage clusters that are more than one kilometer away from \cite{trujillo_fish_2012}'s identified locations. Furthermore,  we identify an additional 8 clusters that postdate their study period -- on aerial imagery from 2013 to 2021 -- and are more than one kilometer away from their identified locations. 
\figurename~\ref{fig:french_facilities} maps the known marine finfish aquaculture locations identified by \cite{trujillo_fish_2012}, and the cage clusters detected by our model during 2000-2021 that are located at least one kilometer away from these known production locations (see the Supplementary Materials for a visualization of these locations over time; \figurename~\ref{fig:french_facilities_time}).

\subsection*{Aquaculture Production Estimation in the French Mediterranean}

\begin{figure}[t!]
    \centering
    \includegraphics[width=0.9\columnwidth]{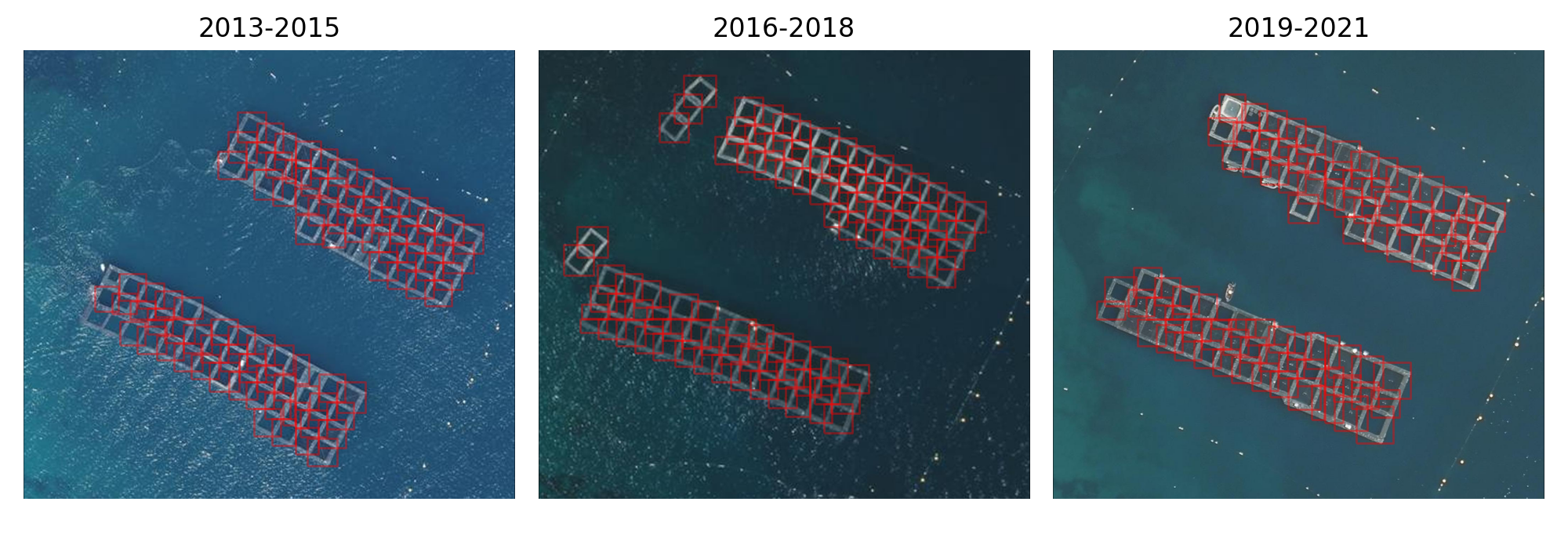}
    \caption{\textbf{Predictions for a cluster of cages in the French Mediterranean over time.}  Imagery: Institut national de l'information géographique et forestière (IGN)}
    \label{fig:facility_detections}
\end{figure}

Our method allows us to directly estimate the number of aquaculture farms at a given point in time as well as number and surface area of cages at each location (\figurename~\ref{fig:facility_detections}).
One downstream analysis these predictions can be particularly useful for is developing a bottom-up estimate of finish production that is almost independent of self-reported and often incomplete survey-based estimates.
To illustrate this application, we compute annualized estimates of finfish marine aquaculture production in the French Mediterranean for the 2000-2021 period using our method, and compare these estimates to FAO statistics reported for the same species, region of interest and time period. 
The aerial imagery along the coast is captured at a cadence of approximately 2-4 years, and at varying frequency for different regions (see the Supplementary Materials for a visualization of the imagery's spatial coverage; \figurename~\ref{fig:French_Mediterranean}).
For this reason, we compute production estimates by grouping the aerial imagery from different years into specific time periods (\textit{e.g.}, 2010, 2011 and 2012, as visualized in \figurename~\ref{fig:French_Mediterranean}) such that they represent a complete view of the entire French Mediterranean coast (we estimate annualized production over each wave of imagery on the assumption that production is relatively stable over this time frame). To produce comparable FAO statistics, we compute the average annual tonnage for each of these time periods.

\begin{figure}[t]
    \centering
    \includegraphics[width=0.8\columnwidth]{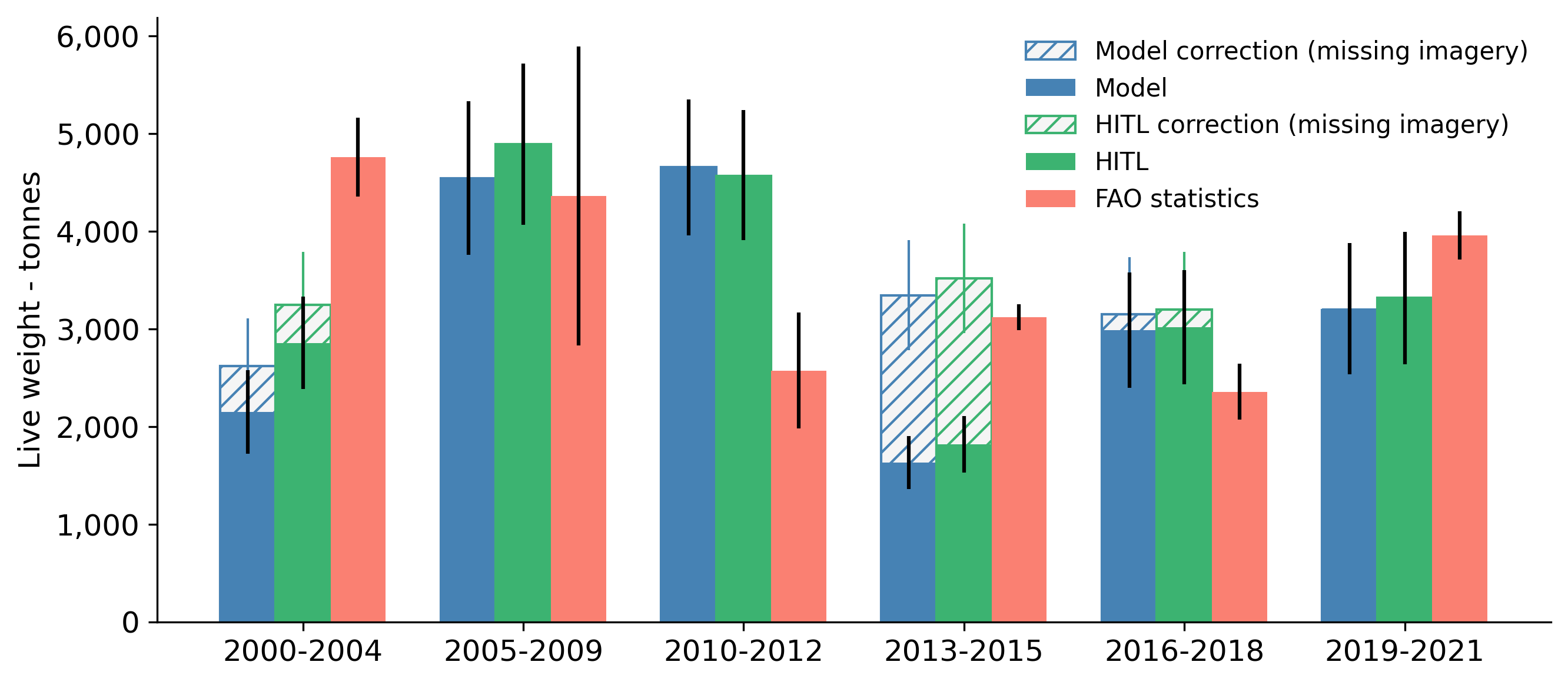}
    \caption{\textbf{Marine finfish aquaculture tonnage in the French Mediterranean over time.} Model estimates reflect annualized tonnage computed from the area of our predicted cages, with error bars reflecting bootstrap standard errors that incorporate uncertainty in the aerial imagery, in model performance and in the tonnage production factors. Human-in-the-loop (HITL) estimates reflect annualized tonnage computed from the area of annotated cages, with error bars reflecting bootstrap standard errors that incorporate uncertainty in the aerial imagery and in the tonnage production factors. Food and Agriculture Organization (FAO) data reflects average annual production reported to FAO during the period, with error bars reflecting the standard deviation of these values. We include estimates of annualized production that account for the fact that aerial imagery is at times unavailable for some locations.}
    \label{fig:tonnage_estimates}
\end{figure}

\figurename~\ref{fig:tonnage_estimates} visualizes the annualized production values (measured in tonnes -- live weight equivalent) estimated by our method and the average annual production values reported by FAO for each time period. In the case of our model estimates, the error bars reflect bootstrap standard errors that account for the following sources of uncertainty: \textit{(1)} model performance; \textit{(2)} differing cage area estimates arising from multiple aerial images for a given location; \textit{(3)} cage area uncertainty given a lack of knowledge of the underlying cage orientation within each detected bounding box; and \textit{(4)} uncertainty in other factors of production (cage depth, stocking density and harvest frequency), modeled according to the distributions described in the Supplementary Materials (\figurename~\ref{fig:factor_distributions}). For FAO statistics, error bars reflect the standard error of the annual production statistics that fall in each time period. Due to aerial imagery being unavailable for some locations in the 2000-2004, 2013-2015 and 2016-2018 time periods, we also include estimates of annualized production that account for this missing imagery. This is accomplished by measuring the tonnage output from these locations using the imagery from a different time period with more complete imagery. Specifically, we use the 2005-2009 period to impute the tonnage from locations with missing imagery for the 2000-2004 period, and use the 2010-2012 period -- which has relatively good spatial coverage -- to impute missing tonnage for both the 2013-2015 and the 2016-2018 periods.

Overall, we find that our tonnage estimates match up quite closely to FAO statistics in most periods, with two exceptions. In 2000-2004, our estimates are significantly lower, driven by lower quality of the coastal imagery that is available for these years, limiting the ability of our model to identify cages. On the other hand, our estimates are significantly higher in 2010-2012. The stark decline in French finfish tonnage exhibited during 2005-2009 -- likely driven by weakened demand at the time of the economic crisis \cite{globefish_european_2009} and reflected in \figurename~\ref{fig:tonnage_estimates} in the wide error bars for the FAO data in this period -- could be an explanation for this discrepancy. It is possible that retired cages are not immediately removed from farms, such that our estimates for the period include the estimated tonnage of a large number of cages that became inactive during this time of rapid production decline.
In addition to FAO statistics, we can compare our tonnage estimates to those of \cite{trujillo_fish_2012}, who also use the cage areas -- derived from their manual survey -- to estimate production. Within one kilometer of the authors' known facilities, we estimate annualized finfish production of 1,890 tonnes during the 2005-2009 period, closely tracking their estimate of 2,008 tonnes produced in 2006 (assuming 75\% of cages are in production).

While the combination of the model predictions and post-processing steps can provide a rapid and robust means of identifying aquaculture locations and measuring their extent and production, one option for improving further on the accuracy of these measurements is to use our method as a means of identifying candidate locations for human review. Such a human-in-the-loop (HITL) approach takes advantage of the domain expertise and judgement that benefits manual surveys of aerial imagery, while significantly reducing the burden of these efforts. For instance, starting from our method's predictions along the coastline, human reviewers would only need to look at 2.8\% of coastal imagery to confirm all predictions. In this manner, the HITL approach combines the model's capacity to signal locations that have a high likelihood of containing cages, and human involvement in verifying the model's predictions. Since the most laborious component of manually searching the coast is precisely finding these rare locations within the vast amount imagery, this framework represents a highly cost-effective and accurate alternative to existing manual efforts. 
\figurename~\ref{fig:tonnage_estimates} also visualizes tonnage estimates computed from the set of manually labeled cage annotations that was derived from examining only the images in which the model detected a cage (with any confidence score), images near known aquaculture production locations, and a small sample of images that neither have predictions nor are close to known locations. While the HITL estimates are similar to the model-derived estimates in general, reflecting the overall high quality of the model's predictions, HITL estimates are notably higher in the 2000-2004 period where poor image quality limited the model's performance. Nonetheless, these estimates remain considerably below the average FAO-reported tonnage for that period, suggesting that the image quality issues may still pose a challenge for the human reviewers as well as the model. As a separate exercise, we can also use the HITL approach to estimate an upper bound on the total number of cages in the coastal imagery from 2000-2021. We estimate a population of 4,285 cages, including 4,010 cages that were manually annotated from 2.8\% of the imagery, and 275 cages estimated from the remaining, un-annotated portion of the imagery.

\section*{Discussion} \label{sec:discussion}

We have developed a highly adaptable framework that leverages high resolution aerial imagery and object detection to locate marine finfish aquaculture locations and estimate their production over time, as well as computing robust measures of uncertainty on these estimates. Furthermore, we have demonstrated the application of our approach in the context of the French Mediterranean, representing, to our knowledge, the first
deep-learning based mariculture mapping application outside of China. Our model and prediction post-processing procedure demonstrate strong performance (91\% precision and 73\% recall at the cage-level on an independent and representative test set), finding all of the known production sites in the region that have been identified in a prior manual survey by  \cite{trujillo_fish_2012} and discovering additional  locations both within the time period this manual search was conducted and in the years after. Moreover, we have found that our production estimates capture at minimum the annual mariculture finfish tonnage reflected in FAO statistics for the region in all but one time period, potentially due to poor aerial image quality in 2000-2004. 

Our methodology for obtaining information about aquaculture locations has a number of potential applications for environmental monitoring and impact studies. First, it enables the acquisition of data on the number of aquaculture cages and production sites in a region, as demonstrated along the French Mediterranean coast, as well as their locations and sizes.
Such detailed information is  unavailable from common aquaculture data sources, such as the FAO's FishStatJ platform for fishery and aquaculture statistics. These data are highly relevant for understanding the spread and scale of aquaculture production at any given point in time and, when compared inter-temporally, can give insight into industry dynamics, such as changes in industry agglomeration and production practices.
Second, it offers an alternative to survey-based estimates of aquaculture production. Remotely-sensed aquaculture data has the potential to allow experts to estimate production where survey data is missing. Furthermore, where survey-based production estimates are available, remote sensing-derived estimates can be used to verify data derived from surveys, which can rely on producers to self-report production amounts \cite{national_marine_fisheries_service_fisheries_2021}.

\begin{figure}[t!] 
    \centering
    \includegraphics[ width=0.49\columnwidth]{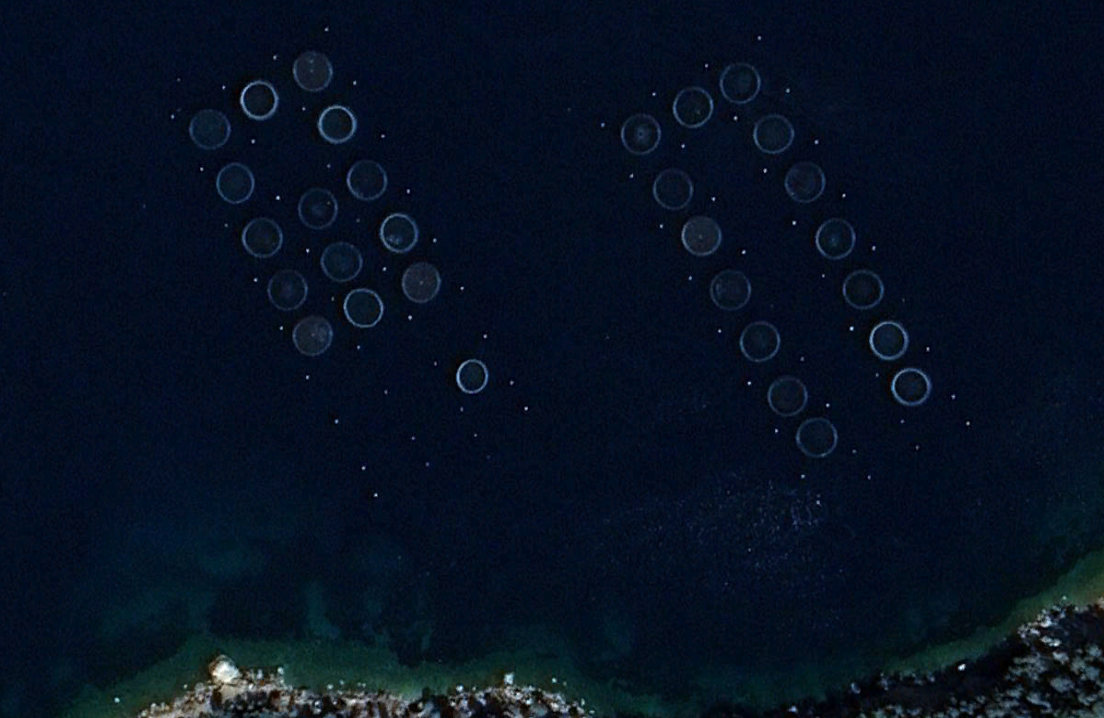}
    \includegraphics[width=0.49\columnwidth]{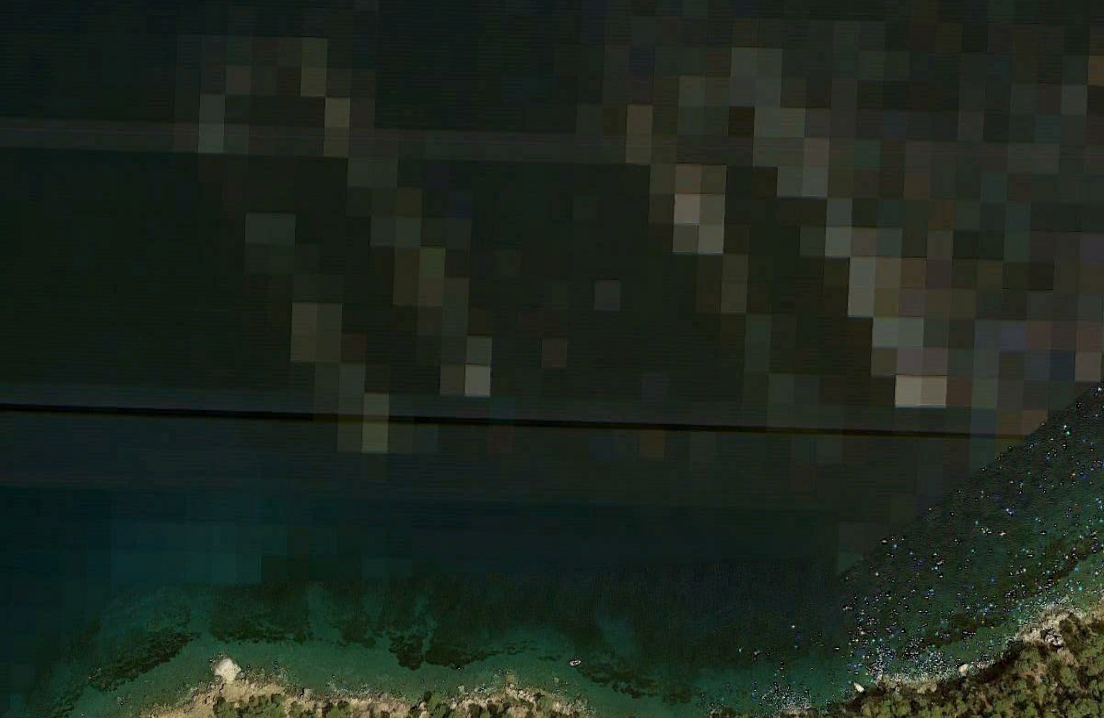} \\
    \caption{\textbf{Pixelated satellite imagery}. Satellite imagery for an aquaculture facility near the Greek island of Poros from April 2019 (left) and June 2021 (right). Imagery: Google, 2023 CNES/Airbus (April 2019 image)}
    \label{fig:blurred_images}
\end{figure}

While our approach offers numerous advantages, we note several limitations to the current work. First, our approach may not generalize to other regions that use different structures for aquaculture production. For instance, our method is unsuitable for detecting underwater aquaculture rafts, which are difficult to identify from aerial imagery. While research interest in submersible cages has increased over the past couple of decades, our method's relevance is underpinned by the fact that surface cage culture remains the dominant production typology, and not all finfish species are well-suited to submersible culture~\cite{sievers_submerged_2022}.
Second, our method is dependent on the availability of high-resolution aerial or satellite imagery, and limited image cadence may limit the granularity of any longitudinal study. In our production estimation of mariculture in the French Mediterranean, the available imagery limited our results to a cadence of 3-4 years over the 2000-2021 period, and lower quality of the imagery from earlier years impacted the performance of our model. The limitations of relying on high-resolution imagery extends to other production systems that our method could potentially be adapted to monitor. For instance, the extraction of inland aquaculture ponds used in freshwater aquaculture -- 44\% of global aquaculture production, of which 57\% is in China \cite{fao_fao_2020} -- typically relies on high-resolution imagery given the narrow dikes that separate these ponds, difficulties distinguishing these from other water bodies, and the complex land cover contexts in which these are located~\cite{zeng_rcsanet_2020,zeng_extracting_2019,han_dynamic_2023}.
The availability of high-resolution imagery of global coastlines and seas varies significantly across time and geography, but continued improvements in the availability, cadence and quality of satellite data~\cite{burke_using_2021} suggest that these kinds of tools are likely to become more useful over time. Separately, we noticed several instances of aquaculture farms that were pixelated to the point of obfuscation in our images. Some of these instances, such as the imagery of an aquaculture facility near the Greek island of Poros (\figurename~\ref{fig:blurred_images}), have arisen after community-guided efforts have pointed to the adverse environmental effects of these farms~\cite{noauthor_petition_2020}.  
This kind of alteration to remote sensing imagery can impact the usefulness of imagery-based methods for environmental monitoring.
Third, there are inherent limitations to measuring the magnitude of aquaculture tonnage with a high level of precision using this object detection on aerial imagery approach. For instance, our area computations are necessarily based on bounding box predictions for individual cages rather than a precise outline of the  cage objects. Although we account for uncertainty stemming from the imprecision of bounding boxes in our estimates, cage predictions from an image segmentation may result in greater precision. Additionally, our method does not identify the extent to which cages are utilized relative to their capacity, which means our estimates may better reflect available production capacity than actual production during periods of low utilization. 
Fourth, even with perfectly accurate cage predictions, computing production tonnage from these annotations is a challenging task due to the need for alternative data sources (\textit{e.g.}, bathymetry for cage depth, stocking densities, harvest frequencies, etc.), an understanding of local regulation and enforcement, and the domain expertise required to model the production factor distributions in order to calculate sound uncertainty measures. Although our uncertainty estimates failed to model more complex relationships such as the correlation of production factors across facilities, we note that our framework can be easily adapted to incorporate more complex mechanisms of this nature.

Although our current study focuses on mariculture in the French Mediterranean as a proof of concept, we believe that our framework can be adapted to other settings. 
With a small amount of training data capturing the dominant cage typologies used for production in this location, and quick fine-tuning of a pre-trained detection model, our approach enabled the comprehensive estimation of the region's mariculture production. In the same manner, we expect that our approach can generalize and be readily implemented in new locations by collecting a modest amount of labels that are representative of the region's cage farms, and re-training the model with this data. Our aim is that the highly adaptive nature of our approach -- which differs from previous studies applying deep learning methods to measure aquaculture production from remote sensing imagery -- can lower barriers to using deep learning for the environmental monitoring of aquaculture and make these tools more accessible to researchers with less experience using deep learning methods. 
Moreover, we have illustrated the use of a separate framework under which computer vision models can be employed with a higher level of human involvement, which we expect could also be readily adapted to new locations. This human-in-the-loop approach efficiently removes the most laborious component of existing manual surveys -- finding the rare locations with aquaculture facilities -- with the highly precise detections that humans can produce. We exemplified this approach with the manual annotation of fewer than 3\% of the entire set of images that was viewed by our model to produce temporal estimates for the entire French Mediterranean coast during a 21-year period.  

To conclude, in this study we developed a method to detect aquaculture production locations by fine-tuning a pre-trained object detection model with a relatively small aquaculture cage label set,  present the first instance (to our knowledge) of using deep learning to detect mariculture farms outside of China, and demonstrate how our method can be used to detect aquaculture sites in the French Mediterranean. Our aim is that this highly adaptable and accessible approach can drastically improve the capacity of aquaculture researchers and regulators to monitor industry dynamics and detect environmental harms, build public awareness around these issues, and build an evidence base for improvements in regulations that balance needs for food security, environmental impact, and animal welfare.

\section*{Materials and Methods} \label{sec:methods}

The subsections below describe the two key components of our marine aquaculture prediction pipeline: \textit{(1)} the data collection, model selection and model training, and the output post-processing procedures to develop our object detection model; and \textit{(2)} the data collection, model evaluation and statistical methods to estimate marine finfish aquaculture in the French Mediterranean using our model. In the Supplementary Materials, we include additional details describing the cage annotation protocol used to build the training and evaluation datasets, the method used to evaluate model performance, the approach used to tune our prediction post-processing procedure, our method to estimate cage areas from bounding box predictions (see Table \ref{tab:area_calculations}), and our tonnage uncertainty quantification methodology (see \figurename~\ref{fig:factor_distributions}).  

\subsection*{Training a computer vision model to detect finfish cages}

\subsubsection*{Training data collection}

To obtain training data for our computer vision model, we identified Mediterranean marine aquaculture sites using a dataset of 1,020 locations manually assembled by \cite{trujillo_fish_2012}. The facilities in this dataset are primarily located in the coastal waters of Greece, Italy, Croatia, France, Albania, and Malta. We utilized Google Earth Pro's (GEP) historical imagery feature to download large, high-resolution images at each location for all available time snapshots. To pre-process the raw images, we downscaled them by 50\% and tiled them into 1024x1024 pixel images. Finally, we randomly selected 893 of these images and manually created bounding box labels around any cages using Hasty.ai (https://hasty.cloudfactory.com), a platform that offers annotation tools for computer vision tasks (see the Supplementary Materials for annotation guidelines). Each cage bounding box was also classified as denoting a \textit{circular cage},  \textit{square cage} or \textit{other cage type}, depending on the cage geometry. In total, our annotated data comprised 1,775 circular cages, 689 square cages and 7 labels belonging to other cage types.

Once our images were labeled, we randomly partitioned them into training, validation and test sets for model training. We created these splits such that 64\% of the cage labels were in the train set, while 16\% and 20\% were in the validation and test set, respectively. This resulted in a partition of 59\%/20\%/21\% of the images into the train, validation, and test sets.

By sampling images around known mariculture facility locations we were able to quickly assemble a dataset containing a large number of positive mariculture cage instances. However, the resulting dataset is highly biased toward locations that are amenable to mariculture production. A model trained on this data may perform poorly when asked to predict on a set of imagery that represents the complete distribution of coastal areas where mariculture activity is sought to be detected. To show that our model performs well in a realistic inference setting, we evaluated its capacity to detect mariculture cages along the entirety of the French Mediterranean coast. 

\subsubsection*{Model architecture and training}

We used an object detection approach to locate mariculture cages from satellite and aerial imagery, which generates bounding boxes around instances of objects that it identities in an image. In contrast, previous studies using neural networks for aquaculture mapping have largely focused on image segmentation models (\textit{e.g.}, \cite{fu2019finer,Fu2022,Shi2018,zou2022extraction}), which predict what class each pixel of an image belongs to. We opted for the object detection approach for its ease-of-use and adaptability, which is mainly illustrated in two parts of the training procedure. 
First, when labeling images with many small, geometrically complex features, such as mariculture cages, it can be faster to create bounding box labels than labels that perfectly resemble the object geometries. In other words, we expect that our model choice reduced the time cost of data collection. Second, there are a vast number of resources that enable practitioners to quickly and easily train well-performing object detection models, including platforms like YOLOv5 and off-the-shelf models that can be adapted to specific detection tasks. 

In terms of model selection, we chose to fine-tune the medium-sized version of the YOLOv5 object detection architecture (yolov5m.pt)~\cite{Jocher_YOLOv5_by_Ultralytics_2020}. Some models, such as YOLOv5, have been pre-trained on large databases of images with bounding box labels (in our case, the MS COCO dataset~\cite{lin_microsoft_2014}). Fine-tuning a pre-trained model involves optimizing pre-trained model weights to generate accurate predictions on additional data that represent a more specific prediction task. 
Fine-tuning models that have been trained on a diverse dataset of images and labels can lead to better model performance on a specific task, as opposed to training an entirely new model from scratch on the task without the use of pre-trained weights, for two reasons. 
First, models trained to generate accurate predictions across a diverse set of images encode information that is relevant for many object detection tasks, which can lead to better performance~\cite{han2021pre}. 
Second, by virtue of having encoded information that is useful across object detections tasks, pre-trained models tend to require less data to achieve strong performance on a task than if the model was trained on randomly instantiated weights. YOLOv5 specifically has been widely used for a range of object detection tasks since its release in 2020 \cite{wu2021application,fang2021accurate,kasper2021detecting}.

To train the model, we fine-tuned YOLOv5 for 50 epochs to generate predictions for three cage typologies (circular, square and other) on 640x640 pixel images (YOLOv5 automatically resizes the 1024x1024 training images to this size), using a batch size of 16 and the default hyperparameters defined for YOLOv5. We use the validation and test sets comprising the GEP historical imagery to monitor the model's performance and guide its training procedure, although we ultimately evaluate its performance on the French Mediterranean coastal imagery.

\subsubsection*{Prediction post-processing}

We implemented two post-processing steps to improve the quality of our model predictions. First, we removed predictions that were located on land by 
constructing a shapefile of French landmass and filtering any predictions that intersected this geometry (see the Supplemental Materials for details on how the French landmass shapefile was constructed). 
Second, we employed the DBSCAN clustering algorithm~\cite{ester1996density} to aggregate our aquaculture cage predictions into cage clusters. DBSCAN represents each aquaculture cage as a node and defines edges between nodes that are within some distance of one another. The resulting connected components are the output clusters. We used the DBSCAN implementation from the sci-kit-learn package (version 0.0.post4)~\cite{scikit-learn}, which also allows users to filter clusters under a given size. As most aquaculture facilities operate a number of aquaculture cages in close proximity, filtering predictions that do not form a cluster of a given size is effective at removing false positive predictions. We tuned all hyperparameters related to output post-processing (the minimum cluster size and distance threshold used by the DBSCAN algorithm) and the selection of a model confidence threshold through a grid search procedure described in the Supplementary Materials.  

\subsection*{Estimating marine aquaculture tonnage in the French Mediterranean}
\subsubsection*{Inference data collection}
We obtained imagery of the French Mediterranean coast for the 2000 to 2021 period from Institut national de l'information géographique et forestière (IGN), a French government agency that maintains geographical information. IGN provides high resolution imagery for each French department (one of the country's administrative subdivisions), captured with a cadence of 2-4 years on different years for varying regions within France. The image resolution varies over time, ranging from 50 cm/pixel for most departments prior to 2014, to 15 cm/pixel in later years.

For our task, we collected imagery from IGN's BD ORTHO series~\cite{bd_ortho} along the the French Mediterranean coast, covering nine departments (Pyrénées-Orientales, Aude, Hérault, Gard, Bouches-du-Rhône, Var, Alpes-Maritimes, Haute-Corse and Corse-du-Sud). To do so, we created a shapefile of French Mediterranean coastal waters by intersecting a shapefile of the French Mediterranean sea with a shapefile of Europe's shoreline buffered by 2000m on each side \cite{french_maritime_shapefile,europe_coastline}.  We then tiled our coastal waters shapefile into 200m by 200m squares, and queried the BD ORTHO data portal with each tile to download an image at the tile's location. As the raw images from BD ORTHO are very large (6144x6144 pixels), we further tiled the downloaded images into 1024x1024 pixel squares before feeding them to the model. 

Due to the different capture years on the imagery for each department, we were unable to obtain imagery for the complete French Mediterranean coast in a single year. For this reason, to generate annualized marine aquaculture tonnages, we combined the images from different years into groups (2000-2004, 2005-2009, 2010-2012, 2013-2015, 2016-2018 and 2019-2021) that comprise a spatial coverage of the coast that is as complete as possible. Our tonnage estimation and uncertainty quantification procedure accounted for several artifacts stemming from the irregularity in the aerial imagery, including \textit{(1)} the availability of multiple imagery from different years for a given location; and \textit{(2)} missing imagery for a location within a given group.

\subsubsection*{Model performance evaluation}

After running inference on the French Mediterranean coastal imagery, we evaluated our model's performance by manually annotating a subset of the coastal imagery. To measure model precision, we annotated all of the images with model predictions. To allow for the estimation of model recall, we also annotated all ocean images without predictions that were near known aquaculture production locations, as well as a random sample of ocean images without predictions that were not near known sites. 
Table \ref{table:strata} describes how we partitioned the entire set of French aerial imagery into strata to define which images would be annotated.  
We first classified images according to whether they were fully contained by the French landmass geometry. Next, we partitioned the images that were not wholly contained on French landmass into two groups: those that had model predictions, and those that did not. Finally, for the images that did not have any predictions, we further disaggregated these into images that were approximately within one kilometer of the aquaculture sites identified by \cite{trujillo_fish_2012}, and those that were not. We note that, as distance calculations were performed using the Coordinate Reference System of the imagery, the threshold used to define whether a location was near a known site was in practice approximately 900 meters rather than one kilometer. In the table, we also present the ocean images with predictions disaggregated according to the maximum model score of a prediction within the image.

\begin{table}
\centering
\begin{tabular}{@{}p{0.46\linewidth}p{1.7cm}p{1.7cm}p{2cm}p{1.5cm}l@{}}
\toprule
\textbf{Stratum} & \textbf{Number of images} & \textbf{Sampled images (\%) } & \textbf{Number of predictions} & \textbf{Number of cages} \\
\midrule
Land                                                                & 187,566    & n/a             & 9,515           & 0             \\
Not land, 0 $\leq$ maximum score \textless 0.3                    & 2,203      & 100\%             & 2,243           & 7             \\
Not land, 0.3 $\leq$ maximum score \textless 0.5                  & 4,868      & 100\%             & 5,316           & 16            \\
Not land, 0.5 $\leq$ maximum score \textless 0.8                  & 3,402      & 100\%             & 5,319           & 38            \\
Not land, 0.8 $\leq$ maximum score \textless 1                    & 1,157      & 100\%             & 7,684           & 3,912         \\
Not land, no prediction, near known location     & 6,846      & 100\%             & 0               & 37            \\
Not land, no prediction, not near known location & 783,355    & 1\%               & 0               & 0             \\ \bottomrule
\end{tabular}
\caption{\textbf{Stratification of the French aerial imagery.} Stratification was based on whether the imagery captured any landmass, the presence of model predictions, and the location of the imagery (whether it was near any of the aquaculture locations found by \cite{trujillo_fish_2012}). None of the images in the \textit{Land} stratum were sampled for annotation, as any predictions from this subset were assumed to be incorrect.}
\label{table:strata}
\end{table}

To generate the sample for annotation, we first excluded any images from the \textit{Land} stratum. In this case, no manual annotation was required as we assumed all positive model predictions to be incorrect. Second, we sampled all of the images from the \textit{Not land} strata that contained model predictions. Third, for the ocean images without predictions (\textit{Not land, no prediction}), we sampled all of the images near known locations and ~1\%  of the images that were not near known locations, as the latter stratum contained hundreds of thousands of images. Our prior was that the set of ocean images without predictions and far from known production locations was highly unlikely to contain any aquaculture cages. However, by sampling a small fraction of the images in this stratum, we were able to estimate a conservative bound on the number of cage labels that may be in this group, despite random sampling not yielding any positive instances in these areas. To estimate this bound, we determine how large the proportion of images with cages could be in this stratum such that we could find -- with a reasonable probability -- zero cages in our sample (see the Supplementary Materials for further details).
Our sampled images were labeled by CloudFactory (https://www.cloudfactory.com), a third party data labeling vendor (see the Supplementary Materials for annotation guidelines).

Our performance metrics of interest were model precision and recall. See the Supplementary Materials for additional information on how these metrics were computed and aggregated across strata.

\subsubsection*{Production estimation and uncertainty measurement}

Our bounding box predictions of aquaculture cages allowed us to estimate total aquaculture production over time with some uncertainty, following a similar methodology to that employed in \cite{trujillo_fish_2012}. To do so, we computed an estimate of the annualized production (measured in tonnes -- live weight equivalent) of finfish marine aquaculture in each time period (2000-2004, 2005-2009, 2010-2012, 2013-2015, 2016-2018 and 2019-2021) by aggregating the estimated production of the clusters of cage predictions. Additionally, we compared the prediction-based anualized production estimates to annual FAO aquaculture production data for France in the Mediterranean and Black Sea region, within marine environments, for all finfish species (meagre, seabream, seabass and miscellaneous marine fishes). As we grouped the imagery from each year into periods that represent a survey of the French Mediterranean coast that is as complete as possible given the available imagery, our aggregated measures are estimates of the region's annualized aquaculture production during each period. To compare these period-level estimates to FAO production data, we simply computed the average annual FAO values within each period.

We estimated the annualized tonnage, $Y_{it}$, of a cluster of cages \textit{i} detected in imagery from time period \textit{t} using the following equation:

\begin{equation}
\label{eq:tonnage}
    Y_{it} = A_{it} \times D_{it} \times S_{it} \times H_{it} \times \frac{1}{1000}
\end{equation}

where $A_{it}$ is the estimated total cage area of the cluster (in square meters), $D_{it}$ is the estimated cage depth (in meters), $S_{it}$ is the estimated stocking density (in kilograms per cubic meter),  $H_{it}$ is the estimated annual harvest frequency (e.g., $H_{it}=2$ for fish harvested every 6 months, reflecting that the same cage volume is re-used twice in a given year) and $\frac{1}{1000}$ is a conversion factor from kilograms to tonnes. 

To measure uncertainty in our estimates, we developed a framework that allowed us to propagate uncertainty from each of the four production factors to the final tonnage estimates. This framework enabled us to not only model the uncertainty in each production factor independently, but also to incorporate uncertainty from artifacts of the French Mediterranean coastal imagery and from the computer vision model's performance in order to generate more robust tonnage estimates for each period. We leveraged a bootstrap approach by independently sampling values, for each cluster of cages, for each of the production factors from the distributions defined in the Supplementary Materials, and combining these to compute the cluster's annualized tonnage for the period using Equation~\ref{eq:tonnage}. For each bootstrap sample, we computed period-level tonnage estimates by summing the cluster-level tonnage in each period. We performed this bootstrap sampling procedure 10,000 times, such that our final period-level estimates and error bars reflect the mean tonnage and the standard deviation across bootstrap samples, respectively. 
See the Supplementary Materials for further details and a visualization (\figurename~\ref{fig:factor_distributions}) of the production factor distributions used to generate the bootstrap samples.

\bibliography{aqua_bib}
\bibliographystyle{Science}

\section*{Acknowledgments}
We thank Jennifer Jacquet and Chiara Piroddi for helpful comments and conversations.

\noindent \textbf{Funding: } This research was funded by Stanford Impact Labs, NSF-Access and a couple of anonymous donors.

\noindent\textbf{Author contributions:}

\noindent Conceptualization: SQ, OA, KR, DH

\noindent Methodology: SQ, AV, OA, KR, DH

\noindent Investigation: SQ, AV, OA

\noindent Visualization: SQ, AV, OA

\noindent Supervision: KR, DH

\noindent Writing--original draft: SQ, AV, OA

\noindent Writing--review and editing: SQ, AV, KR, DH

\noindent\textbf{Competing interests:}
The authors declare that they have no competing interests.

\noindent\textbf{Data and materials availability:}
All data and code needed to evaluate the conclusions in the paper are available in the following repository: \verb|github.com/reglab/aquaculture|.\\

\appendix

\clearpage

\newgeometry{textwidth=16cm,textheight=21cm,footskip=1.0cm}
\setlength{\parskip}{\baselineskip}%
\setlength{\parindent}{0pt}%
\singlespacing

\pagenumbering{arabic}
\renewcommand{\thepage}{S\arabic{page}}

\setcounter{equation}{0}
\renewcommand{\theequation}{S\arabic{equation}}

\setcounter{section}{0}
\renewcommand{\thesection}{S\arabic{section}}

\setcounter{table}{0}
\renewcommand{\thetable}{S\arabic{table}}
\setcounter{figure}{0}
\renewcommand{\thefigure}{S\arabic{figure}}
\renewcommand{\figurename}{Fig.}
\renewcommand{\tablename}{Table}
\title{Supplementary Materials for \\
\Large{\textbf{Locating and measuring marine aquaculture production from space: a computer vision approach in the French Mediterranean}}}

% Place the author information here.  Please hand-code the contact
% information and notecalls; do *not* use \footnote commands.  Let the
% author contact information appear immediately below the author names
% as shown.  We would also prefer that you don't change the type-size
% settings shown here.

\author
{Sebastian Quaade \textit{et al.} \\ \\
\normalsize{Corresponding author: Daniel E. Ho,  deho@stanford.edu.}
}

% Include the date command, but leave its argument blank.

\date{}

\maketitle 

\subsection*{This PDF file includes:}
\vspace{-0.3cm}
\begin{adjustwidth}{15mm}{}
Supplementary Text \\
Figs. S1 to S7 \\
Tables S1 to S2 \\
\end{adjustwidth}

\clearpage

\section*{Supplementary Text}

\underline{Cage annotation protocol from imagery} \\
In this section, we describe our procedure to annotate cages from aerial and satellite imagery using bounding boxes, in order to create the datasets used to train the object detection model and to evaluate model performance in the French Mediterranean. While we generated the dataset for model training by following this protocol ourselves, to generate the French Mediterranean dataset we employed CloudFactory -- an IT services company that offers human-in-the-loop solutions for machine learning and business purposes -- as our third-party annotator. Given an aerial or satellite image as input, the task consisted of two steps:
\vspace{-3mm}
\begin{enumerate}
    \item Annotators first had to identify whether any marine finfish cages were present in the image and, if any cages were present, to determine the correct cage type (square or circular). To aid annotators in this identification process, we developed a flow chart (\figurename~\ref{fig:annot_flowchart}) that begins by determining whether a body of water is present in the image and whether physical structures can be observed within this water body, and outlines the physical characteristics of these structures that can be used to determine whether they are in fact square or circular cages. In cases where it was challenging to determine whether an object was a marine finfish cage, annotators were asked to use Google Earth Pro to inspect the location, querying for imagery that was as close as possible to the date of the provided image.
    \item Annotators had to draw individual bounding boxes around each identified marine finfish cage. In order to compute precise area estimates and uncertainty measures, these bounding boxes were required to be as tight as possible around each square and circular cage.
\end{enumerate}
\vspace{-3mm}
We provided annotators with several examples of images for different steps along the flow chart. Additionally, we offered guidance on the annotation procedure for images showing more difficult annotation cases, such as the presence of non-finfish rectangular cages, the presence of cages on land, and the presence on cages in lower-quality imagery. \figurename~\ref{fig:annot_examples} visualizes a subset of these examples.

\underline{Constructing a French landmass shapefile} \\
We filter false positive model predictions on land by constructing a shapefile representing French landmass and removing any predictions that intersect with its geometry. To do so, we perform a series of geometric manipulations using a coarse 10km x 10km resolution outline of French territory (including naval territory)~\cite{france_shapefile} and a high resolution shapefile of the European coastline from the European Environment Agency~\cite{europe_coastline}, as well as a coarse shapefile of French marine territory from Flanders Marine Institute~\cite{french_maritime_shapefile}. First, we take the difference of the polygon representing the entire French territory and the polygon representing French marine areas to obtain a coarse outline of French landmass. Next, we dissect the coarse French landmass geometry with the high resolution outline of the European coast, yielding disjoint polygons whose boundaries are high resolution representations of the locations where French land meets French ocean. We inspect these polygons manually, and combine those that correspond to land into a single multipolygon.

%\vspace{5mm}
\underline{Evaluating model performance} \\
We characterized the performance of our model by measuring model precision and estimating model recall using the stratified sample of annotated images. 
Let $I$ be the population of images on which labels $Y$ and predictions $\hat{Y}$ reside. We partitioned the set of images into the $n=7$ strata defined in Table \ref{table:strata}, which we denote as $I_1, \dots, I_7$. We annotated all of the images in $I_1, ..., I_6$, and a random sample $S_7$ of the images from $I_7$. We note that although we did not directly annotate the images in $I_1$, the \textit{Land} stratum, we assumed that this stratum did not contain any marine finfish cages, and so all model predictions falling within this set of images were marked as incorrect.   
For any set of images $J$, let $Y_J$ and $\hat{Y}_J$ be the labels and predictions on the images, respectively.

To measure precision and estimate recall, our quantities of interest were the empirical ratio of true positive predictions to the total number of predictions and to the total number of labels, respectively. Let $u$ be the bounding box of a model prediction (label) and let $V$ be the set of bounding boxes belonging to a set of labels (predictions). We defined a true positive instance as any time the bounding box of a label and prediction overlapped, as described in the following equation:  

\begin{equation}
\label{eq:true_positive}
    TP(u; V)  := 
    \begin{cases}
        1 &\quad\text{if}\quad \sum_{v\in V} \mathbbm{1} [u \cap v] > 0 \\
        0 &\quad\text{otherwise} 
    
    \end{cases}
\end{equation}

We measure precision and recall both at the cage and cage cluster-level. In the case of cages, we defined whether an instance was a true positive using the detected or annotated bounding boxes for the cage.
In the case of cage clusters, we defined whether an instance was a true positive using the bounding box given by the union of all cage-level detected or annotated  bounding boxes belonging to a cluster. For both cages and cage clusters, we emphasize that we only considered
overlap between predictions and labels  belonging to the same image -- and thus, belonging to the imagery from the same \textit{year} -- to define a true positive instance.

To measure cage-level and cage cluster-level precision we simply computed the ratio of true positive predictions to total predictions, given that we annotated all of the images that had model predictions:

\begin{equation}
    PR = \frac{1}{|\hat{Y}|} \sum_{\hat{y}\in \hat{Y}} TP(\hat{y}; Y)
\end{equation}

We computed cage-level and cage cluster-level recall as follows. As we annotated all of the images belonging to $I_1, ..., I_6$, we estimated population-level recall by measuring recall in $I_1, ..., I_6$ and estimating recall in $I_7$ using the sample of images $S_7$. We used the number of labels in each stratum as weights to aggregate these estimates to the population level, as follows:
\begin{align}
    \widehat{RE} &= \frac{|Y_{I_{1-6}}|}{|Y|} RE_{1-6} + \frac{|Y_{S_7}|}{|Y|} \widehat{RE}_7 \nonumber \\
    &= \frac{|Y_{I_{1-6}}|}{|Y|} \sum_{y \in Y_{I_{1-6}}} \frac{1}{|Y_{I_{1-6}}|} TP(y; \hat{Y}) + \frac{|Y_{S_7}|}{|Y|} \sum_{y \in Y_{S_7}} \frac{1}{|Y_{S_7}|} TP(y; \hat{Y})
\end{align}

Note that since $|Y_{S_7}| = 0$ (see Table \ref{table:strata}), it follows that $|Y_{I_{1-6}}| = |Y|$ and so our estimate of recall is further simplified to the following equation:
\begin{align}
    \widehat{RE} &= 
    \frac{1}{|Y|} \sum_{y \in Y} TP(y; \hat{Y})
\end{align}

\underline{Estimating an upper bound on the population of aquaculture cages} \\
Our stratum including images with some ocean that have no predictions and are more than one kilometer away from a location belonging to the dataset of mariculture sites of \cite{trujillo_fish_2012} -- the \textit{Not land, no prediction, not near known location} stratum in Table \ref{table:strata} -- contained more images than we were able to manually inspect for aquaculture cages. We randomly sampled ~1\% of these images for manual inspection and found no aquaculture cages on them. However, the small size of the sample relative to the stratum population meant there was a non negligible likelihood that we drew a random sample with no aquaculture cages, despite their presence in the population. As a result, the population of aquaculture cages may have actually been larger than what we found from fully sampling the other strata.

We estimate an upper bound on the number of cages in this stratum as follows. First, we estimate how large the proportion of images with cages could be in this stratum such that we could find, with a 50\% probability, zero cages in a sample of images of size $m = 10,518$ -- the number of images that we actually sampled from this stratum. By drawing $10,000$ binomial samples of size $m$ for each proportion $p \in \{0.000010, 0.000011, ..., 0.00009, 0.0001\}$, we find that the proportion of images that have cages could be no larger that $p=0.00007$ in order to have a sample with zero labels, with a 50\% probability. With this proportion, we estimate an upper bound on the number of cages in the \textit{Not land, no prediction, not near known location} stratum as $p \times |I_7| \times k$ cages, where $k = 5$ is the average number of cages per image.

\underline{Tuning of the detection post-processing procedure} \\
We tuned the model score threshold, as well as the maximum distance and minimum cluster size parameter for the DBSCAN algorithm using a broad grid search. We used a $k$-fold cross-validation approach in which we partitioned all of the images in our data that were classified as not belonging to land into five folds. Per this approach, we used each fold as a tuning dataset and the remaining folds as a training dataset. 
We post-processed the predictions in the tuning set using each possible combination of the following parameter values, and then estimated precision and recall with the resulting predictions on the true labels in the tuning set:

\begin{equation}
    CT = \{0.600, 0.605, 0.610, ..., 0.995, 1.000 \}
\end{equation}
\begin{equation}
    DT = \{10, 30, 50,, ..., 130, 150 \}
\end{equation}
\begin{equation}
    MC = \{1, 2, 3, 4, 5, 6, 7, 8, 9, 10 \}
\end{equation}

where $CT$ is the set of explored confidence thresholds, $DT$ is the set of explored distance thresholds (in meters), and $MC$ is the set of explored minimum cluster sizes. 
To select the optimal combination of post-processing parameters, we chose the combination that maximized the product of precision and recall of the post-processed predictions. For robustness, we also explored using the combination that maximized the F1 score, and found that both metrics retrieved the same hyperparameter choices.

Finally, to characterize their true performance, we computed the precision and recall of model predictions in the test dataset after post-processing using the selected hyperparameters.

\underline{Cage area calculations from bounding boxes} \\
In this section, we outline the computation of the area estimate, minimum area and maximum area of each cage from a bounding box of width $w$ and height $h$ (in meters), according to the cage type (square or circular). Given that the model's detected bounding boxes and the annotated bounding boxes need not be perfectly square, we compute the underlying cage area for circular cage predictions using area calculations for ellipses, and compute the underlying cage area for square cage predictions using area calculations for rectangles. The area calculations for each cage type are summarized in Table \ref{tab:area_calculations} and visualized in Fig.~\ref{fig:area_calcs}.

For circular cages that are not on the border of an image, we assume that the height and width of the bounding box are equal to the length of the ellipse's major and minor axes. The surface area of a circular cage is then given by:

\begin{equation}
    A_{circular} = \pi (w/2) (h/2) = \pi w h / 4
\end{equation}

As the bounding box width and height perfectly identify the ellipse area, these estimates have zero uncertainty, and so the maximum and minimum cage area are equal to the central estimate.

However, bounding boxes of circular cages that are not fully contained in an image do not necessarily identify the ellipses' principal axes. In these instances, we assume that true cage surface area is uniformly distributed between the minimum and maximum surface area of a partial ellipse that is bounded by the bounding box. When a bounding box is in the corner of an image, the minimum possible surface area occurs when the bounding captures an ellipse chord that approximates a diagonal from one corner of the bounding box to the other; the maximum possible surface area occurs when the bounding box captures a perfect quarter ellipse. When a bounding box is located on one edge of an image, the minimum possible surface area occurs when the ellipse approximates a triangle circumscribed in the bounding box, and the maximum possible surface area occurs when the ellipse approximates the rectangular shape of the bounding box. In both of these cases, the bounds of the distribution of a cage's true surface area are given by:

\begin{equation}
    A^{max}_{circular} = \frac{\pi wh}{4}
\end{equation}
\begin{equation}
    A^{min}_{circular} = \frac{wh}{2}
\end{equation}

We then derive the central estimate using the properties of a uniform distribution:

\begin{equation}
    A_{circular} = \frac{(A_{max} + A_{min})}{2}
\end{equation}

We also treat the orientation of a square cage within a bounding box as a random draw from a uniform distribution. In this case, the minimum possible surface area occurs when the square cage's vertices touch the midpoint of the bounding box's edges. On the other hand, the maximum area occurs when the square cage is equivalent to the bounding box. 

\begin{equation}
    A^{max}_{square} = wh
\end{equation}
\begin{equation}
    A^{min}_{square} = \frac{wh}{2}
\end{equation}

Since we assume that the rotation angle of the underlying cage within a bounding box is uniform, the central estimate is then given by:

\begin{equation}
    A_{square} = \frac{3}{4}wh = \frac{1}{2} \times \frac{wh}{2} + \frac{1}{2} \times wh
\end{equation}

\underline{Production factor distributions for tonnage uncertainty quantification} \\
For a cluster of cages \textit{i} detected in imagery from time period \textit{t}, we defined the distribution of each production factor to estimate the cluster's tonnage $Y_{it}$ and generate bootstrap samples. Fig.~\ref{fig:factor_distributions} visualizes example distributions of each of the four production factors (cage area, cage depth, stocking density and harvest frequency), which were defined as follows:

\textit{Cage area} \\
Cage area samples were drawn from the following Uniform distribution:

\begin{equation}
\label{eq:area_uniform}
    A_{it} \sim U(A^{min}_{it}, A^{max}_{it})
\end{equation}

where $A^{min}_{it}$ is the minimum total cage area of cluster \textit{i} from time period \textit{t} and $A^{max}_{it}$ is the maximum. The stochasticity for the cage area combined three sources of uncertainty: \textit{(1)} model uncertainty regarding the cage predictions; \textit{(2)} uncertainty regarding the true cage area, given that we capture rectangular bounding boxes around the cages using our detection model; and \textit{(3)} uncertainty in the image selection for locations that have multiple imagery from different years due to overlap in the spatial coverage of the annual IGN imagery. Each of these sources contributed to defining the bounds of the Uniform distribution from Equation \ref{eq:area_uniform}. Let $j$ be a cage prediction from cage cluster $i$ from time period $t$ of type $c \in \{square, circular\}$ that is output by our method (\textit{i.e.}, by the detection model and the tuned prediction post-processing procedure). Then, let $A_{cjit}$ be the estimated area of the underlying cage, computed from the width and height of the output bounding box as outlined in Table \ref{tab:area_calculations}.

First, to incorporate model prediction error, we began by estimating the distributions of the model errors using our cage predictions and the annotations from the stratified sample. We stratified our predictions according to the cage type $c$ and period $t$. Then, for each cage $j$ predicted by our model, we found the annotated cage $k$ of type $c$ from period $t$ with the highest spatial overlap, and computed the error as the difference between the area of the annotated cage and that of the predicted cage. For each cage type and time period, we used these errors to fit a Normal distribution, $ N(\mu_{ct}, \sigma_{ct})$. We fit 12 Normal distributions in total, reflecting the 2 cage types and 6 time periods. Once we computed these error distributions, we adjusted the estimated area of each cage $j$ of type $c$ from period $t$ as follows:

\begin{equation}
    \Tilde{A}_{cjit} = A_{cjit} + \epsilon_{cjit}, \text{ where } \epsilon_{cjit} \sim N(\mu_{ct}, \sigma_{ct})
\end{equation}

We ensured that $\Tilde{A}_{cjit} > 0$ by sampling the errors until this condition was met for all cages. 

Second, to capture uncertainty in the actual surface cage area, we leveraged our classification of cage predictions as circular or square to obtain estimates of the minimum and maximum possible underlying cage area for each prediction. From the adjusted cage area $\Tilde{A}_{cjit}$, we obtained the upper and lower bounds reflecting the maximum and minimum cage area, $\Tilde{A}_{cjit}^{min}$ and $\Tilde{A}_{cjit}^{max}$, depending on the cage type $c$, using the computation outlined in Table \ref{tab:area_calculations}. From this step, we obtained an aggregate upper and lower bound on the cage area of cage cluster \textit{i} from time period \textit{t} as follows:

\begin{equation}
    \Tilde{A}_{it}^{min} = \sum_{j \in \text{ cluster } i} \Tilde{A}_{cjit}^{min} \\
\end{equation}
\begin{equation}
    \Tilde{A}_{it}^{max} = \sum_{j \in \text{ cluster } i} \Tilde{A}_{cjit}^{max} \\
\end{equation}

Third, to capture uncertainty in the image selection within time periods, we began by optimizing the selection of the imagery prior to bootstrapping. To do so, we selected the combination of all available images from different years that result in the maximum and minimum cage area for each location along the coast (see Fig.~\ref{fig:French_Mediterranean} for an example of the spatial coverage of the aerial imagery of the French Mediterranean within a time period). This image selection defines which specific cage predictions are incorporated into the tonnage estimates. Let $s \in \{image max, image min\}$ denote this image selection scheme. For each image selection scheme, we separately computed the upper and lower bounds on the aggregate cage area for the cluster, $\Tilde{A}_{cjit}^{min}(s)$ and $\Tilde{A}_{cjit}^{max}(s)$. It is important to note that to ensure that the adjusted cage area under selection scheme $s=imagemin$ was less or equal to that of $s=imagemax$, we sampled errors at the cage-level, such that the same cage suffers the same model error adjustment under both schemes. 

Finally, we combine the minimum and maximum aggregate cage areas under each image selection scheme to compute the upper and lower bounds on the Uniform distribution from Equation \ref{eq:area_uniform} from which we will sample the cage area for cluster $i$:

\begin{equation}
    A_{it}^{min} = \Tilde{A}_{it}^{min} (s=imagemin)
\end{equation}
\begin{equation}
    A_{it}^{max} = \Tilde{A}_{it}^{max} (s=imagemax) \\
\end{equation}

%\vspace{5mm}
\textit{Cage depth} \\
Cage depth samples were drawn from a mixture of two truncated Normal distributions:
\begin{equation}
\label{eq:cage_depth_distribution}
    D_{it} \sim \kappa TN(\mu_l = \hat{d}_i,\sigma_l = \sigma_{li}, a_l = 1, b_l  =\hat{d}_i) + (1-\kappa) TN(\mu_r = \hat{d}_i, \sigma_r = \sigma_{ri}, a_r = \hat{d}_i, b_r = 2\hat{d}_i)
\end{equation}

where $\kappa \in [0, 1]$ defines the mixture component weights, $\mu_{z}$ defines the Truncated Normal mean, $\sigma_{z}$ defines the Truncated normal standard deviation, and $a_{z}$ and $b_{z}$ define the lower and upper bounds of the Truncated Normal. We define these for $z \in \{l, r\}$, the Truncated Normal components on the left-hand and right-hand side of $\hat{d}_i$, our estimate of the cage-depth for a cluster of cages $i$. 

We modeled this distribution according to the bathymetry data for each cluster of cages and according to FAO cage aquaculture guidelines. These guidelines recommend that the cage depth should be no larger than half of the water depth \cite{cardia_guidelines_2017}. While the guidelines are not mandatory, facilities have strong incentives to maximize the cage depth within a close neighborhood of this threshold due to the risk of cage abrasion when water levels are low. Cages are tethered to the ocean floor such that they rise and lower with the change in water levels, which implies that cage damage could occur at sufficiently low water levels if the cage depth is too high. The mixture of Truncated Normal distributions models this optimization problem faced by facilities: cage depth should ideally be as close to the threshold as possible to maximize production, and cage depth can indeed go above the recommended threshold but not too far above due to potential risks of abrasion. The use of a mixture in the distribution for cage depth allowed us to better model this asymmetry in the probability mass of the cage depth.

To define the distribution in Equation \ref{eq:cage_depth_distribution} for cluster $i$, we first computed an estimate of the water depth at the location of the cluster, $z_i$, using 2022 bathymetry data from the European Marine Observation and Data Network~\cite{european_marine_observation_and_data_network_emodnet_emodnet_2022}. As the resolution of the bathymetry data is somewhat coarse (115m x 115m), and some cage clusters may be very close to shore, we estimated $z_i$ as the maximum bathymetry value observed within the cage detections belonging to a cluster. Where bathymetry data was unavailable (this was the case for 21 out of 136 cage clusters), we used an average cage depth estimate of 4.84m, which was derived from \cite{trujillo_fish_2012} as the average cage depth for France.  Additionally, we imposed a minimum threshold for the cage depth of 1 meter ($\Bar{d} = 1$), to be conservative in our uncertainty measures. We used $z_i$ to estimate the cage cluster-level cage depth recommended by FAO, $\hat{d}_i$ as follows:

\begin{equation}
    \hat{d}_i = \begin{cases}
        \max\{\frac{z_i}{2}, \Bar{d}\} \text{ if } z_i \text{ is available at cluster }i \\
        4.84 \text{ if } z_i \text{ is unavailable at cluster }i
    \end{cases}
\end{equation}

Then, the mixture of Truncated Normal distributions from Equation \ref{eq:cage_depth_distribution} defines a cage depth distribution over the interval between $a_l$, the minimum cage depth (assumed to be 1 meter), and $b_r = 2 \times \hat{d}_i$, the estimated water depth at the cluster. 
For both truncated Normal distributions, we chose the standard deviation for a cluster to be $\sigma_{li}= \sigma_{ri} = \frac{b-a}{1.96}$. This models the spread of the Truncated Normal distribution to be proportional to that of a Normal distribution Z such that $P(Z < a_l) = 0.025$ in the case of the left-hand Truncated Normal, and such that $P(Z > b_r) = 0.025$ in the case of the right-hand Truncated Normal.  
We chose $\kappa=0.5$.

%\vspace{5mm}
\textit{Stocking density} \\
Stocking density samples were drawn from the following truncated Normal distribution:
\begin{equation}
    S_{it} \sim TN(\mu=\mu_t, \sigma=\sigma_t, a=5, b=20)
\end{equation}

where $\mu$ and $\sigma$ are the mean and standard deviation of the Truncated Normal, and $a$ and $b$ are its lower and upper bounds, respectively. The upper bound of the truncated Normal was set to $a=20kg/m^3$, according to the technical rules that oversee mariculture activities in each French department. This is the maximum average stocking density established in the technical rules of most departments that have finfish aquaculture production activities (\textit{e.g.}, Var \cite{direction_departementale_des_territoires_et_de_la_mer_du_var_arrete_2020} and Bouches-du-Rhône \cite{direction_departementale_des_territoires_et_de_la_mer_arrete_nodate}). 
On the other hand, the lower bound of the truncated Normal, $b$, was based on literature describing commercial aquaculture practices \cite{noauthor_animal_nodate}.  

For the mean and standard deviation of this distribution, we inferred parameter values from the grey literature for three finfish species: seabass, seabream and meagre. These species represent most of France's finfish marine aquaculture production, accounting for 83.2\% of the country's marine finfish production in the Mediterranean in 2021 \cite{fao_fao_2020}. The remaining 16.8\% of marine finfish aquaculture production in 2021 was classified in FAO statistics under ``marine fishes nei" or miscellaneous marine fishes; to compute the parameter values using the production shares, we re-classified the tonnage from this category as seabream. The mean and standard deviations for each species derived from the literature are summarized in Table \ref{tab:factor_parameters}.

We aggregated these species-level parameters into period-level stocking density and harvest frequency parameter values using the annual share of aquaculture production across species as weights. Species-level production statistics for each time period were derived from FishStatJ, FAO's platform for fishery and aquaculture statistics \cite{fao_fao_2020}. Let $X_{s,t}$ be the marine production of species $s$ in time period $t$ (in tonnes), $X_t$ be the total marine finish production in time period $t$. Then, we estimated the mean and variance of production factor $\rho_{t}$ at time period $t$ from the species-level factors $\rho_{s, t}$ as follows:

\begin{equation}
    \Bar{\rho_t} = \sum_{s\in S} \frac{X_{s,t}}{X_t} \rho_{s,t}
\end{equation}

\begin{equation}
    Var(\Bar{\rho_t}) = \sum_{s\in S} (\frac{X_{s,t}}{X_t})^2  \times Var(\rho_{s,t})
\end{equation}

\textit{Harvest frequency} \\
Annual harvest frequency samples were drawn from the following Normal distribution:

\begin{equation}
    H_{it} \sim N(\mu=\mu_t, \sigma=\sigma_t)
\end{equation}

As in the case of the stocking density, $\mu_t$ and $\sigma_t$ were computed as the weighted average of the harvest frequency and harvest frequency standard deviation, respectively, of each species according to the production weights of each species in a given time period.

\pagebreak 

\begin{figure}[h]
    \centering
    \includegraphics[ width=1\columnwidth]{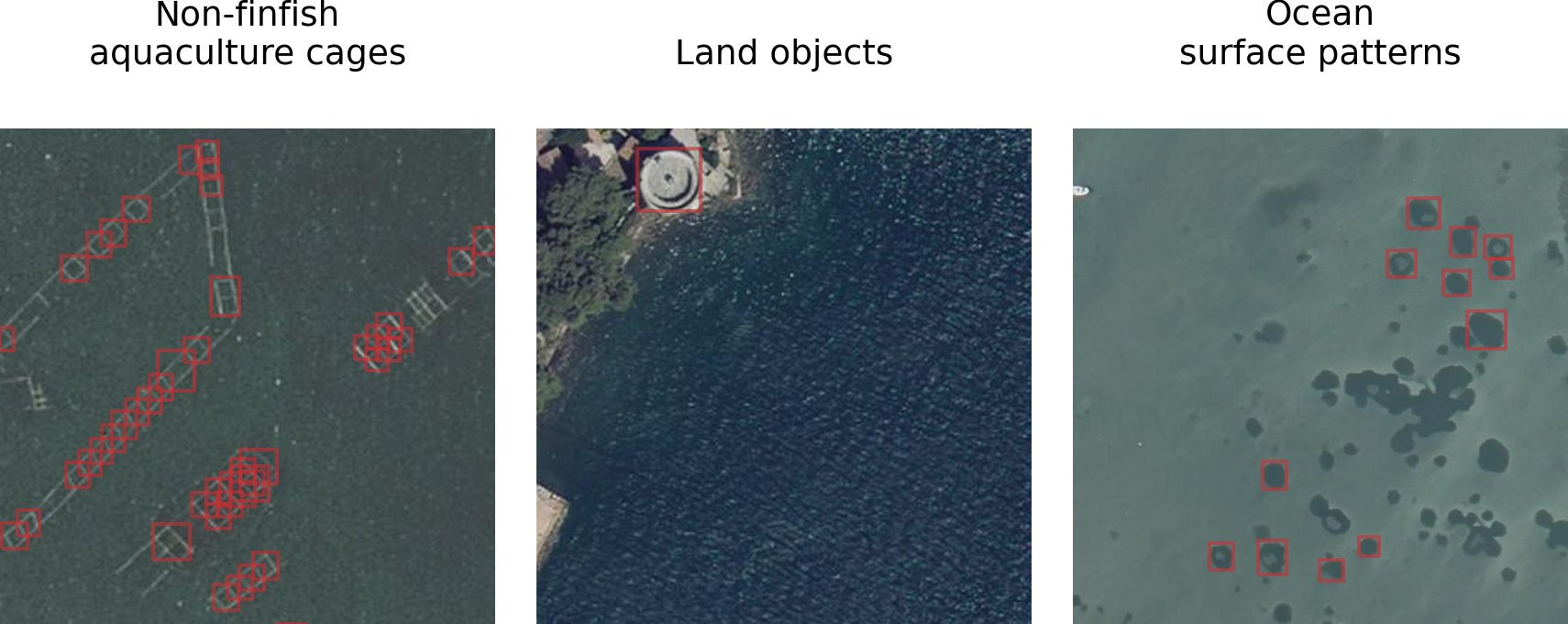}
    \caption{\textbf{Examples of common false positive predictions from the detection model.} Common false positive predictions (red bounding boxes) from the detection model in the French Mediterranean aerial imagery include non-finfish aquaculture cages (\textit{e.g.}, shellfish rafts or other aquatic raft structures), circular structures on water and land, and other water-based objects such as boats and boat yards. The two post-processing steps (land filtering and cage clustering) that complement our detection model improve overall precision through the removal of some of these false positive instances. Imagery: Institut national de l'information géographique et forestière (IGN) }
    \label{fig:inference_false_positives}
\end{figure}

\section*{Fig. \ref{fig:inference_false_positives}}

\pagebreak 

\begin{figure}[ht!]
    \centering
    \includegraphics[ width=1\columnwidth]{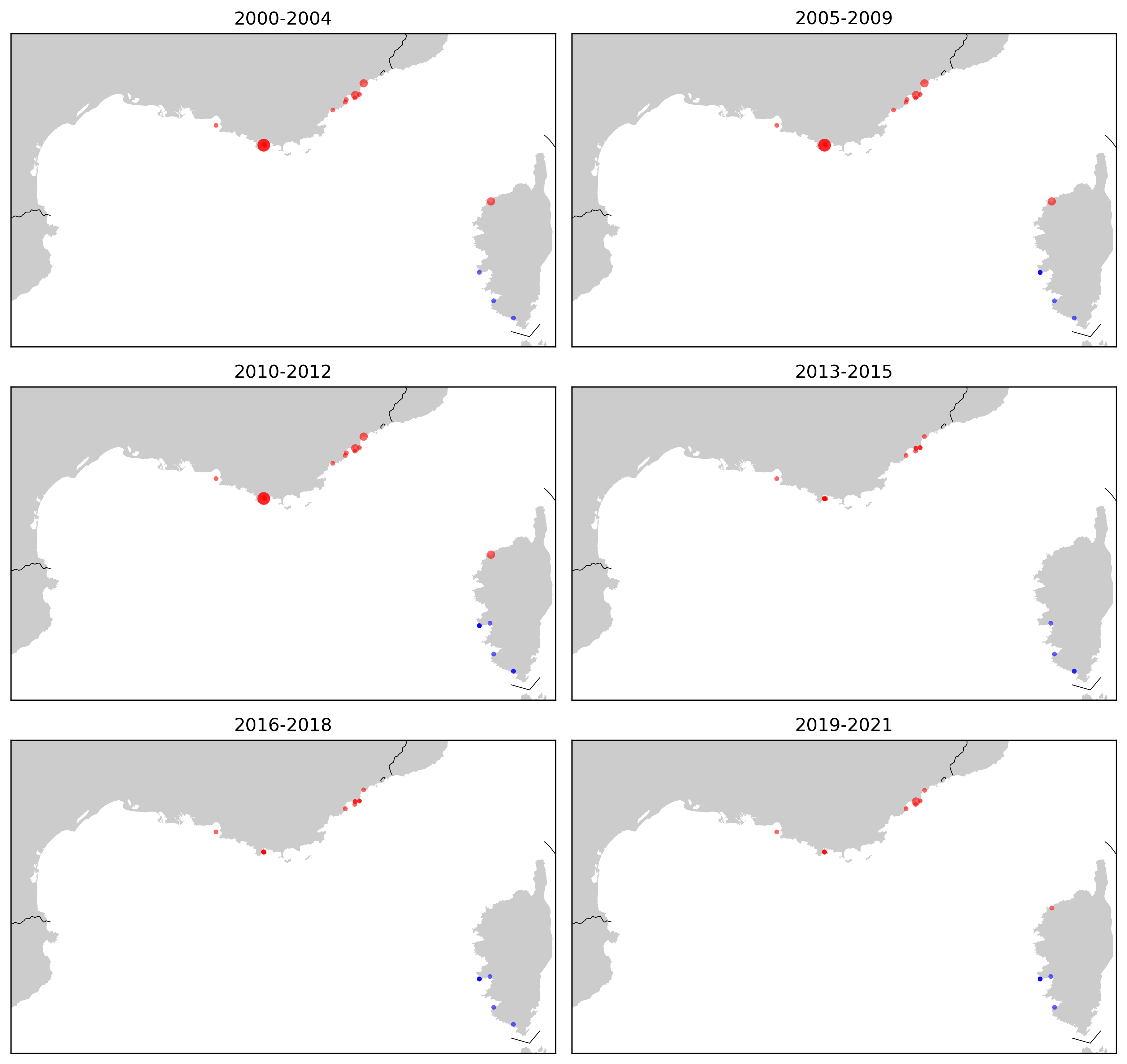}
    \includegraphics[width=0.5\columnwidth]{MapsLegend.pdf} \\
    \caption{\textbf{Marine finfish aquaculture production locations in the French Mediterranean.}  Red points indicate the known locations found by \cite{trujillo_fish_2012} in their manual survey of 2002-2010 Google Earth. 
    Blue points indicate cage clusters detected by our model that are at least one kilometer away from these known locations.}
    \label{fig:french_facilities_time}
\end{figure}

\section*{Fig. \ref{fig:french_facilities_time}}

\pagebreak

\begin{figure}[ht]
    \centering
    \includegraphics[width=\columnwidth]{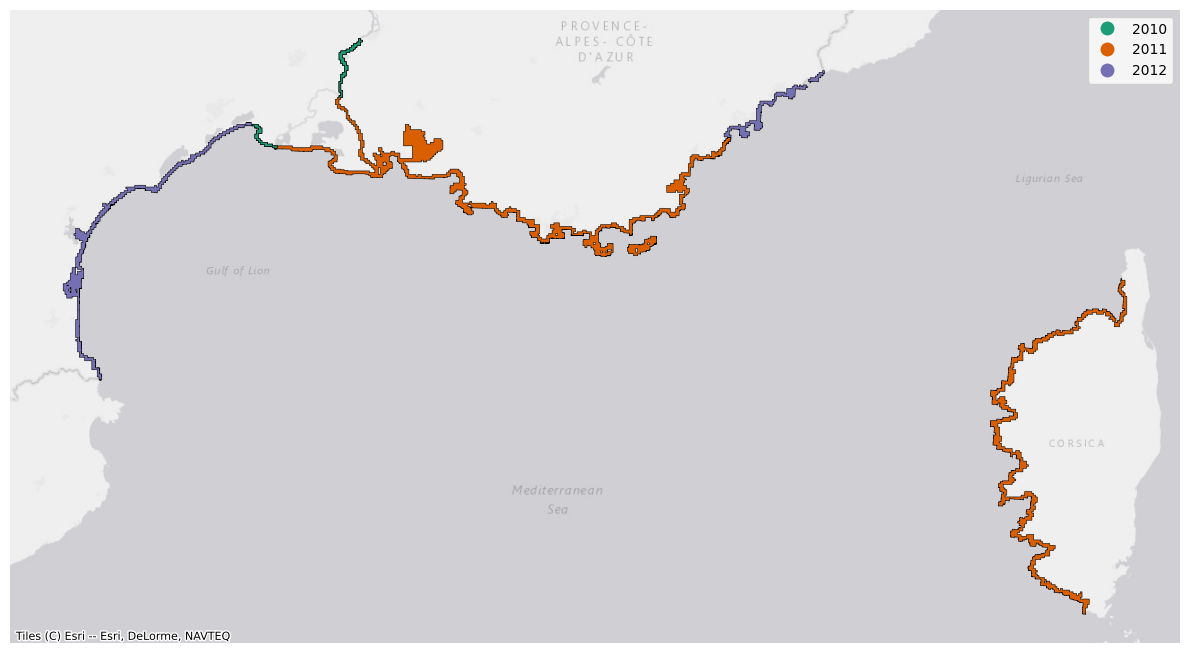}
    \caption{\textbf{Spatial coverage of the aerial imagery of the French Mediterranean.} The imagery -- provided by Institut national de l'information géographique et forestière (IGN) --  at each location is captured at a cadence of approximately 2-4 years, and is captured in different years for different regions along the coast. The figure shows the spatial coverage belonging to 2010, 2011 and 2012, whose imagery we combine to generate annual aquaculture production estimates for the 2010-2012 period for the entire French Mediterranean.   }
    \label{fig:French_Mediterranean}
\end{figure}

\section*{Fig. \ref{fig:French_Mediterranean}}

\begin{figure}[ht!]
    \centering
    \includegraphics[width=\columnwidth]{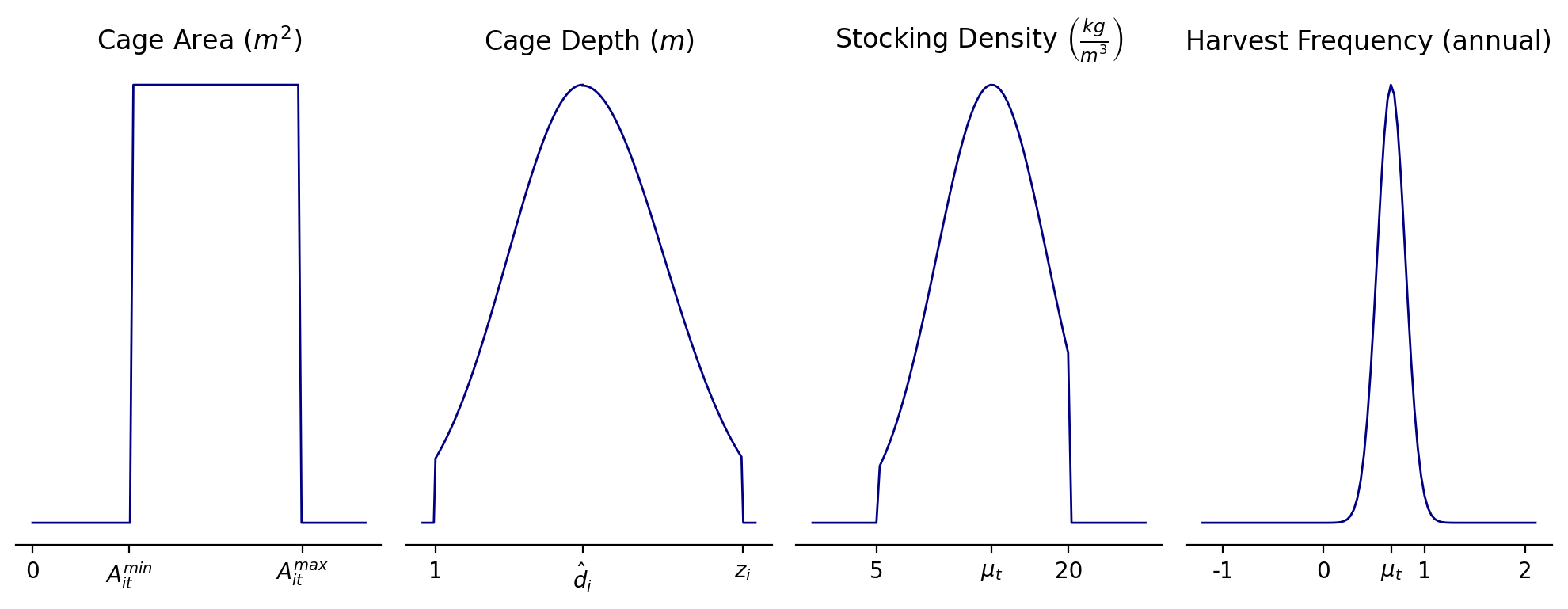}
    \caption{\textbf{Example distributions of the production factors used in tonnage estimation.} The distributions for each of the four production factors used to compute finfish aquaculture tonnage estimates and uncertainty measures were designed according to bathymetry constraints, FAO guidelines for cage culture, French regulation on mariculture practices and finfish life cycles.}
    \label{fig:factor_distributions}
\end{figure}

\pagebreak
\section*{Fig. \ref{fig:factor_distributions}}

\begin{figure}[ht!]
    \centering
    \includegraphics[width=0.65\columnwidth]{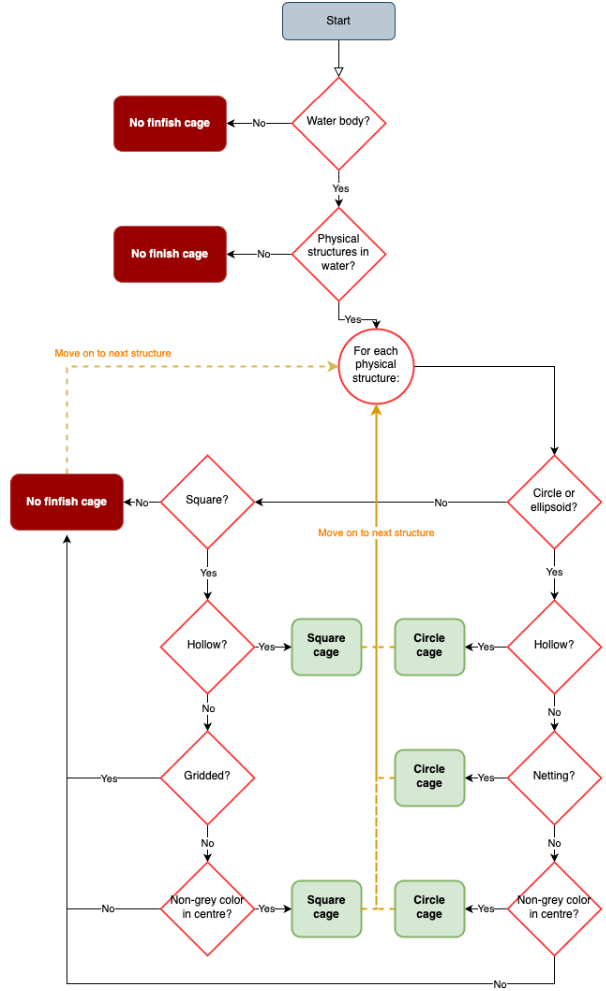}
    \caption{\textbf{Annotation protocol to identify marine finfish cages.} Given an aerial or satellite image, annotators used this protocol to determine whether there were any marine finfish cages in the image, and what type of cage structure (square or circular) these represented.}
    \label{fig:annot_flowchart}
\end{figure}

\pagebreak
\section*{Fig. \ref{fig:annot_flowchart}}

\begin{figure}[ht!]
    \centering
    \begin{tabular}{c c c}
        No water body & Contains water body & Hollow circular structures\\
        \includegraphics[ width=0.25\columnwidth]{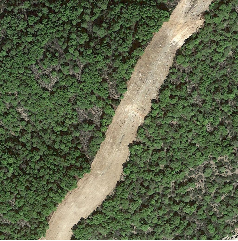} & 
        \includegraphics[width=0.25\columnwidth]{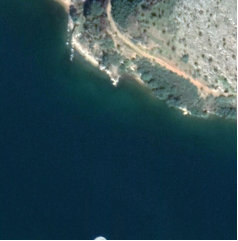} 
        & \includegraphics[width=0.25\columnwidth]{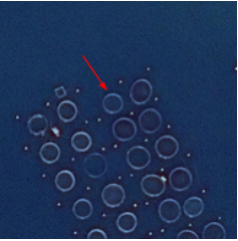} \\
        Cage with a broken net & Cages with netting & Feed/storage platforms \\
        \includegraphics[ width=0.25\columnwidth]{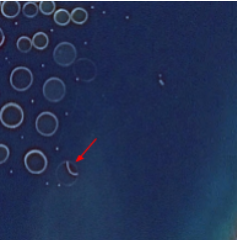} & 
        \includegraphics[width=0.25\columnwidth]{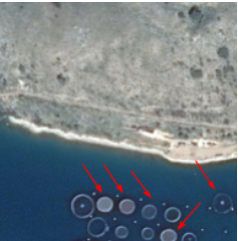} 
        & \includegraphics[width=0.25\columnwidth]{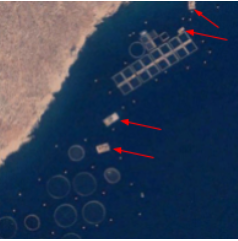} \\
        Cages in low-quality image & Cage on land & Non-finfish cages \\
        \includegraphics[ width=0.25\columnwidth]{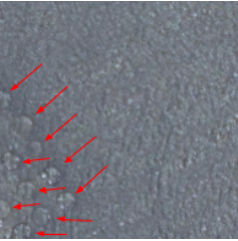} & 
        \includegraphics[width=0.25\columnwidth]{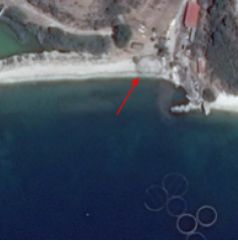} 
        & \includegraphics[width=0.25\columnwidth]{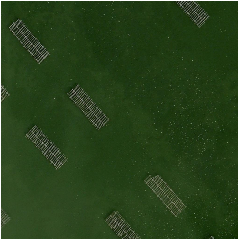}
    \end{tabular}
\caption{\textbf{Examples provided in the annotation protocol}. Annotators were provided examples of different images containing land, water bodies, and finfish cages, as well as difficult edge cases such as images containing non-finfish, rectangular cages. Imagery: Google, 2023 CNES/Airbus}
\label{fig:annot_examples}
\end{figure}

\pagebreak
\section*{Fig. \ref{fig:annot_examples}}

\begin{figure}[ht]
    \centering
    \includegraphics[width=\columnwidth]{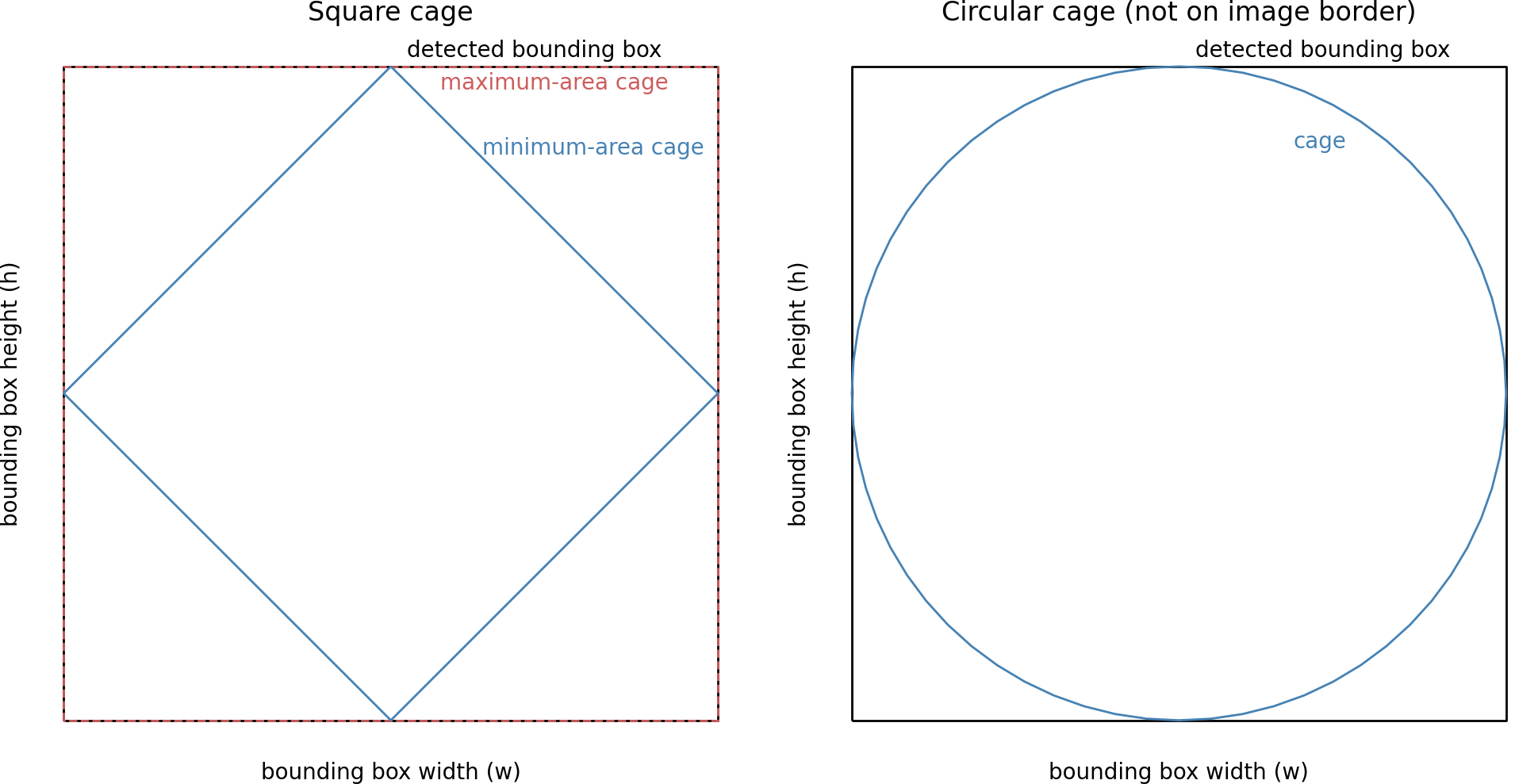}
    \caption{\textbf{Calculation of the underlying cage area from rectangular bounding boxes.} Our cage area estimates from square or circular cage detections reflect uncertainty in the underlying cage rotation within a bounding box. }
    \label{fig:area_calcs}
\end{figure}

\pagebreak
\section*{Fig. \ref{fig:area_calcs}}

\pagebreak
\section*{Table \ref{tab:area_calculations}}
\begin{table}[ht]
\centering
\begin{tabular}{p{4.2cm}p{3.4cm}p{3.4cm}p{3.4cm}}
\toprule       
\textbf{Cage type ($c$)}  & \textbf{Maximum cage area} & \textbf{Minimum cage area} & \textbf{Area estimate}   \\
 & ($A_c^{max}$) & ($A_c^{min}$) & ($A_c$)  \\
\midrule
\textbf{Circular} &  &  &  \\
\rule{0pt}{3ex} \textit{Not on the image border} & $\frac{\pi w h}{4}$ & $\frac{\pi w h}{4}$ & $\frac{\pi w h}{4}$  \\
\rule{0pt}{3ex} \textit{On the image border} & $\frac{\pi wh}{4}$ & $\frac{wh}{2}$ & $\frac{1}{2} \times (A_c^{max} + A_c^{min})$ \\
 \midrule
\textbf{Square} & $wh$ & $\frac{wh}{2}$ & $\frac{3wh}{4}$  \\
\bottomrule
\end{tabular}%
\caption{\textbf{Cage area computation from model predictions or cage annotations.} Cage area estimate, minimum cage area and maximum cage area estimates from the height ($h$) and width ($w$) of each bounding box (in meters) according to the cage type and location of the bounding box within the imagery.}
\label{tab:area_calculations}
\end{table}

\pagebreak 
\section*{Table \ref{tab:factor_parameters}}
\begin{table}[ht]
\centering
\begin{tabular}{p{4.9cm}p{1.9cm}p{1.7cm}p{1.5cm}p{1.8cm}p{1.7cm}}
\toprule       
\textbf{Factor} and \textit{Species}  & \textbf{Trujillo et al. \cite{trujillo_fish_2012} estimate} & \textbf{Estimate} & \textbf{Range} & \textbf{Standard deviation} & \textbf{Source}  \\
\midrule
\textbf{Stocking density} ($kg/m^3$)&  &  &   \\
\textit{ Meagre} & NA & 15.00 & 10-15 & 1.44 & \cite{monfort_present_2010}\\
\textit{ Sea bass} & 12.00 & 20.00 & 5-20 & 4.33 & \cite{noauthor_animal_nodate} \\
\textit{ Sea bream} & 12.00 & 12.50 & 5-20 & 4.33 & \cite{noauthor_animal_nodate, francois_finfish_2010}\\
 \midrule
\textbf{Harvest frequency} \textit{(annual)} & &  &   \\
\textit{ Meagre} & NA & 0.67 & 0.5-1.0 & 0.14 & \cite{francois_finfish_2010, kruzic_meagre_2016}\\
\textit{ Sea bass} & 0.67 & 0.60 & 0.5-1.2 & 0.20 & \cite{garcia_garcia_life_2019, zoli_life_2023, abdou_rearing_2018} \\
\textit{ Sea bream} & 0.75 & 0.67 & 0.5-1.2 & 0.20 & \cite{garcia_garcia_life_2019, zoli_life_2023, abdou_rearing_2018}\\
\bottomrule
\end{tabular}%
\caption{\textbf{Estimates and uncertainty measures for species-level stocking density and harvest frequency.} We include \cite{trujillo_fish_2012}'s estimates for comparison. Standard deviations were derived for the estimate assuming a uniform distribution over the parameter range found in the literature.}
\label{tab:factor_parameters}
\end{table}

\end{document}